\documentclass[10pt,letterpaper]{mystyle}
\usepackage{natbib}

\usepackage{wrapfig}
\usepackage{graphicx}
\usepackage{hyperref}
\usepackage{url}
\usepackage{enumitem}
\usepackage{subcaption}
\usepackage{caption}
\usepackage{booktabs}
\usepackage{wrapfig}
\usepackage{multicol}
\usepackage[nameinlink]{cleveref}
\usepackage{todonotes}
\usepackage{multicol}
\usepackage{multirow,makecell}
\graphicspath{{./plots/}}
\title{CoPeP: Benchmarking Continual Pretraining for Protein Language Models}
\runningtitle{CoPeP: Benchmarking Continual Pretraining for Protein Language Models}

\author[1,2,4]{Darshan Patil}
\author[1,3,4]{Pranshu Malviya}
\author[1,2,4]{Mathieu Reymond}
\author[1,2]{Quentin Fournier}
\author[1,3,4,5]{\\Sarath Chandar}

\affil[1]{Mila - Quebec AI Institute}
\affil[2]{University of Montréal}
\affil[2]{Polytechnique Montréal}
\affil[4]{Chandar Research Lab}
\affil[5]{Canada CIFAR AI Chair}

\correspondingauthor{Darshan Patil \href{mailto:darshan.patil@mila.quebec}{darshan.patil@mila.quebec} and Quentin Fournier \href{mailto:quentin.fournier@mila.quebec}{quentin.fournier@mila.quebec}}

\newcommand{\ft}{\mathcal{F}_t}
\newcommand{\dt}{\mathcal{D}_t}
\newcommand{\di}[1]{\mathcal{D}_{#1}}

\begin{abstract}
    Protein language models (pLMs) have recently gained significant attention for their 
    ability to uncover relationships between sequence, structure, and function from 
    evolutionary statistics, thereby accelerating therapeutic drug discovery.
    These models learn from large protein databases that are continuously updated by the 
    biology community and whose dynamic nature motivates the application of continual 
    learning, not only to keep up with the ever-growing data, but also as an opportunity 
    to take advantage of the temporal meta-information that is created during this 
    process.
    As a result, we introduce the Continual Pretraining of Protein Language Models 
    (CoPeP) benchmark, a novel benchmark for evaluating continual learning approaches on 
    pLMs.
    Specifically, we curate a sequence of protein datasets derived from the UniProt 
    Knowledgebase spanning a decade and define metrics to assess pLM performance across 
    31 protein understanding tasks.
    We evaluate several methods from the continual learning literature, including replay, 
    unlearning, and plasticity-based methods, some of which have never been applied to 
    models and data of this scale.
    Our findings reveal that incorporating temporal meta-information improves perplexity 
    by up to 7\% even when compared to training on data from all tasks jointly.
    Moreover, even at scale, several continual learning methods outperform naive 
    continual pretraining.
    The CoPeP benchmark offers an exciting opportunity to study these methods at scale
    in an impactful real-world application.
\end{abstract}

\begin{document}

\maketitle 
\begin{abstract}
    Protein language models (pLMs) have recently gained significant attention for their 
    ability to uncover relationships between sequence, structure, and function from 
    evolutionary statistics, thereby accelerating therapeutic drug discovery.
    These models learn from large protein databases that are continuously updated by the 
    biology community and whose dynamic nature motivates the application of continual 
    learning, not only to keep up with the ever-growing data, but also as an opportunity 
    to take advantage of the temporal meta-information that is created during this 
    process.
    As a result, we introduce the Continual Pretraining of Protein Language Models 
    (CoPeP) benchmark, a novel benchmark for evaluating continual learning approaches on 
    pLMs.
    Specifically, we curate a sequence of protein datasets derived from the UniProt 
    Knowledgebase spanning a decade and define metrics to assess pLM performance across 
    31 protein understanding tasks.
    We evaluate several methods from the continual learning literature, including replay, 
    unlearning, and plasticity-based methods, some of which have never been applied to 
    models and data of this scale.
    Our findings reveal that incorporating temporal meta-information improves perplexity 
    by up to 7\% even when compared to training on data from all tasks jointly.
    Moreover, even at scale, several continual learning methods outperform naive 
    continual pretraining.
    The CoPeP benchmark offers an exciting opportunity to study these methods at scale
    in an impactful real-world application.
\end{abstract}

\section{Introduction}
\label{sec:introduction}

Proteins are the fundamental building blocks of life, acting as the primary machinery in 
all living organisms.
Their biological functions are largely determined by their three-dimensional structure, 
which is primarily encoded within their linear sequence of amino acids, and by cellular 
context.
Accurately mapping protein sequences to their biophysical properties remains one of the 
core challenges of computational biology.
Recently, protein language models (pLMs) have emerged as an effective and scalable 
solution~\citep{rivesBiologicalStructureFunction2021, linEvolutionaryscalePredictionAtomiclevel2023, madani2023large, nijkampProGen2ExploringBoundaries2023, fournierProteinLanguageModels2024}.
By treating amino acids as ``letters'', functional regions as ``words'', and whole proteins 
as ``sentences'', pLMs leverage evolutionary statistics from large databases to uncover 
relationships between sequence, structure, and 
function~\citep{rivesBiologicalStructureFunction2021, notin2023proteingym}.
Notably, pLMs have demonstrated remarkable accuracy in predicting protein properties and 
even designing novel proteins~\citep{hayes2025simulating}, underscoring their potential 
to significantly accelerate drug discovery.

Despite their effectiveness, pLMs face a significant challenge in the dynamic nature of 
their training data~\citep{fournierProteinLanguageModels2024}.
These models learn from enormous, ever-expanding public databases such as the UniProt 
Knowledgebase~\citep{theuniprotconsortiumUniProtUniversalProtein2025} (UniProtKB), where 
millions of new proteins are deposited each year by the community, while millions of 
others are curated out, either via automated pipelines or by UniProt curators.
Retraining models from scratch on each new data release is inefficient and 
computationally prohibitive.
This challenge, however, also presents a unique opportunity: the temporal evolution of 
these databases provides valuable metadata.
Sequences that persist over time serve as strong examples of valid protein-coding
sequences, whereas those that are later curated out can be treated as implicit examples of
likely non-protein sequences.
By leveraging this history, a model can more effectively learn the language of proteins.
\begin{figure*}
	\centering
	\includegraphics[width=.8\textwidth]{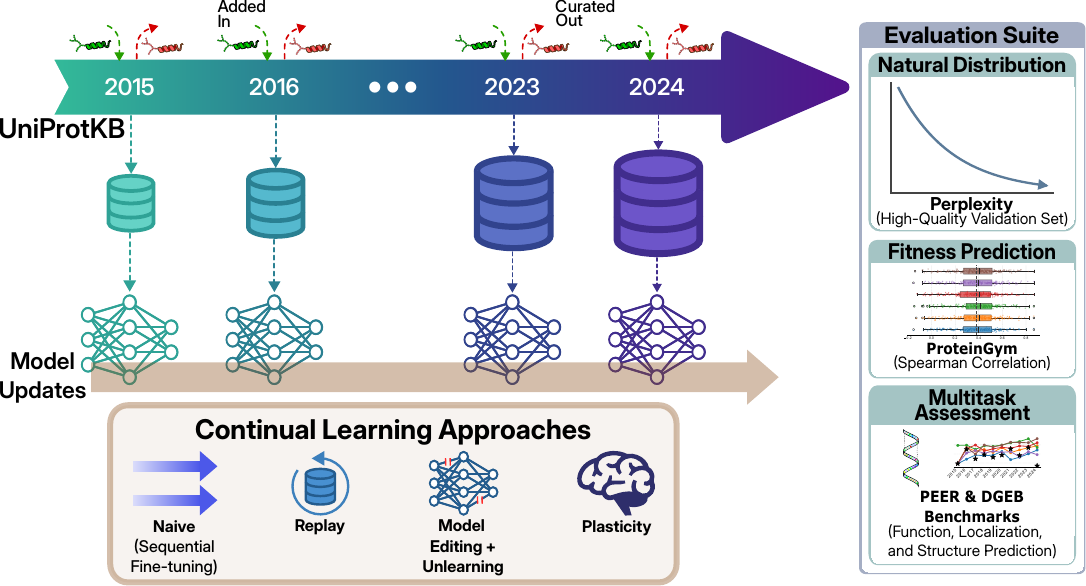}
	\caption{
	Overview of the CoPeP benchmark.
	The UniProt Knowledgebase is continuously updated by biologists,
	which reflects the continuous discovery and curation of proteins.
	For every year's task, we pull the latest UniRef100 release derived from that year's
	UniProtKB release and use it to update the model. We evaluate several continual learning
	approaches on this sequence of datasets and assess their performance on a variety of
	protein understanding tasks.
	}
	\label{fig:cover}
\end{figure*}

Although continual learning is a well-established 
field~\citep{wangComprehensiveSurveyContinual2024} with numerous artificial 
benchmarks~\citep{kirkpatrick2017overcoming, zenke2017continual,
lopezGradientEpisodicMemory2017}, 
there is a growing demand for realistic, large-scale alternatives.
Indeed, while these controlled environments effectively measure loss of plasticity and 
catastrophic forgetting, they fail to reflect the scale and complexity of real-world data.
With the rise of Large Language Models (LLMs), there has been considerable interest in
the community in exploring ways to continually update these models with new information.
Prior works have explored targeted unlearning and editing of memorized 
facts~\citep{bourtouleMachineUnlearning2021, yaoEditingLargeLanguage2023} or investigated 
continual pre-training for domain and language 
expansion~\citep{gururanganDontStopPretraining2020, chalkidisLEGALBERTMuppetsStraight2020, keContinualPretrainingLanguage2022, abbesRevisitingReplayGradient2025}.
However, there remains a lack of datasets and benchmarks designed to study the temporal
evolution of the pre-training distributions themselves.

To bridge this gap, we introduce CoPeP (Continual Pretraining for Protein Language
Models), a realistic, large-scale benchmark designed to evaluate continual learning
approaches for biological sequence modeling.
We leverage the temporal evolution of UniRef100, a non-redundant clustering of UniProtKB 
and the most common source of data for pLMs, by curating a sequence of datasets spanning 
$10$ yearly releases.
We evaluate several popular state-of-the-art methods from the continual learning 
literature, including Gradient Ascent~\citep{golatkarEternalSunshineSpotless2020},
Random Label Learning~\citep{golatkarEternalSunshineSpotless2020}, Hare 
and Tortoise~\citep{leeSlowSteadyWins2024}, Shrink and
Perturb~\citep{ashWarmStartingNeuralNetwork2020}, and
Replay~\citep{rolnickExperienceReplayContinual2019a, chaudhryTinyEpisodicMemories2019a}, 
applying some of them 
for the first time at a scale comparable to real-world applications.
We evaluate our models on three types of tasks: 
(1) a high-quality validation set of experimentally verified proteins to assess 
performance on natural protein distributions~\citep{fournierProteinLanguageModels2024}; 
(2) ProteinGym~\citep{notin2023proteingym}, which measures the ability to predict the 
effects of protein mutations; 
(3) PEER~\citep{xuPEERComprehensiveMultiTask2022} and 
DGEB~\citep{west-robertsDiverseGenomicEmbedding2024}, which focus on predicting various 
aspects of protein function, localization and structure.
Our findings reveal that temporal metadata yields a measurable improvement over models 
trained on individual years, and that several of these methods outperform naive continual
pretraining.
Our contributions are threefold.
First, we introduce CoPeP, a novel benchmark for continual learning on real-world protein 
databases.
Second, we evaluate 7 state-of-the-art continual learning methods on CoPeP, many of
which have never been applied to models and data of this scale.
Third, we demonstrate that temporal metadata contained in the history of proteins added
to or removed from the database can be leveraged to improve the performance of pLMs
beyond that of standard i.i.d. training on individual years.

\section{Related Work}
\label{sec:related_work}

\paragraph{Continual Learning and Model Updating} Continual learning is a machine
learning paradigm in which models are trained incrementally on a sequence of data or
tasks, aiming to accumulate and refine knowledge over time.
Research in this area
primarily focuses on two key challenges: catastrophic forgetting, i.e., the degradation
of previously acquired knowledge~\citep{mccloskey1989catastrophic,
kirkpatrick2017overcoming}, and loss of plasticity, i.e., the diminishing ability to
adapt to new data~\citep{dohare2024loss}.
While some studies investigate continual
learning under natural data shifts~\citep{koh2021wilds, lin2021clear, cai2021online,
bornschein2023nevis}, the datasets involved are typically several orders of magnitude
smaller than modern pretraining corpora: most research relies on academic benchmarks
such as CIFAR-10 and MNIST~\citep{goodfellow2013empirical, zenke2017continual,
krizhevsky2009learning, rebuffi2017icarl}.
While these allow for controlled experimental
setups and the study of severe distribution shifts such as random re-labeling, these
synthetic shifts are often unrealistic, and their small size raises questions about the
generalization of existing methods to larger, more complex scenarios.

Recently, the field has shifted toward updating large pretrained models through model
editing, which updates specific facts~\citep{mengLocatingEditingFactual2022,
mitchellMemoryBasedModelEditing2022}, and model unlearning, which removes the influence
of specific data points~\citep{bourtouleMachineUnlearning2021,
jangKnowledgeUnlearningMitigating2023}.
A few studies have explored continual
fine-tuning, in which a model is iteratively updated across a sequence of downstream
tasks~\citep{jinLearnContinuallyGeneralize2021}, and continual pretraining, in which the
pretraining process itself is extended to incorporate new data.
However, while this has been
explored through domain-adaptive pretraining on distinct, specialized
domains~\citep{gururanganDontStopPretraining2020,
chalkidisLEGALBERTMuppetsStraight2020}, they are typically narrow in scope and their
datasets are small relative to the general pretraining corpora.
A notable exception is
\citet{guptaContinualPreTrainingLarge2023}, who studied the dynamics of pretraining a
large language model on two datasets in sequence.
Nevertheless, practical applications
require methods that scale to longer sequences of datasets, and to our knowledge, no
prior study has considered the temporal evolution of a single large-scale pretraining
dataset.

\paragraph{Protein Language Models} Research in natural language processing (NLP) has
recently been adapted to biology by treating the amino-acid sequence of proteins as a
language.
This perspective has led to the development of protein language models (pLMs),
biology-inspired analogues of NLP models.
For example, the autoregressive
ProGen2~\citep{nijkampProGen2ExploringBoundaries2023,bhatnagarScalingUnlocksBroader2025}
is based on GPT-2~\citep{radfordLanguageModelsAre}, while the bidirectional
ESM~\citep{rivesBiologicalStructureFunction2021,
linEvolutionaryscalePredictionAtomiclevel2023} and
AMPLIFY~\citep{fournierProteinLanguageModels2024} draw inspiration from
BERT~\citep{devlinBERTPretrainingDeep2019}.
Trained on large, diverse, and ever-growing
protein sequence databases~\citep{suzekUniRefClustersComprehensive2015,
jumperHighlyAccurateProtein2021, richardsonMGnifyMicrobiomeSequence2023}, these models
aim to discover the underlying principles that govern protein structure and function by
capturing evolutionary relationships from sequence alone.
This approach has made pLMs an
essential tool in computational biology for a wide range of applications, including
mutational effect prediction, protein structure modeling, and de novo protein
design~\citep{hayes2025simulating}.

Contrary to the holistic benchmarks used in NLP, protein language models are evaluated
on specialized benchmarks that target distinct biological properties.
For example, protein
folding is traditionally measured on the biannual CASP
challenge~\citep{jCriticalAssessmentMethods2018}, in which tertiary structures are inferred
from primary sequences.
On the other hand, functional fitness is primarily evaluated on
the ProteinGym benchmark~\citep{notin2023proteingym}, in which the effects of mutations are
predicted in a zero-shot setting.
Several multi-task benchmarks have been proposed to
evaluate a broader range of concepts such as function, subcellular localization, and
protein-protein interactions.
These include
TAPE~\citep{raoEvaluatingProteinTransfer2019},
PEER~\citep{xuPEERComprehensiveMultiTask2022}, and
DGEB~\citep{west-robertsDiverseGenomicEmbedding2024}.
In this work, we focus on the last
two categories due to their significant role in the drug discovery pipeline and because
protein folding from sequence-only pLMs still lags behind specialized, geometry-aware
architectures like AlphaFold3~\citep{abramsonAccurateStructurePrediction2024}.

\section{CoPeP Benchmark}
\label{sec:benchmark}

To bridge the gap between continual learning research and its practical application, we
introduce CoPeP, the Continual Pretraining for Protein Language Models benchmark.
CoPeP reflects the challenge of keeping protein language models updated with rapidly evolving
biological data.
It serves as a large-scale, complex testbed for evaluating continual
learning methods, with significant implications for protein modeling and drug discovery.

\subsection{Dataset}
\label{subsec:dataset}

\begin{wrapfigure}{r}{0.45\textwidth}

	\centering
	\includegraphics[width=.45\textwidth]{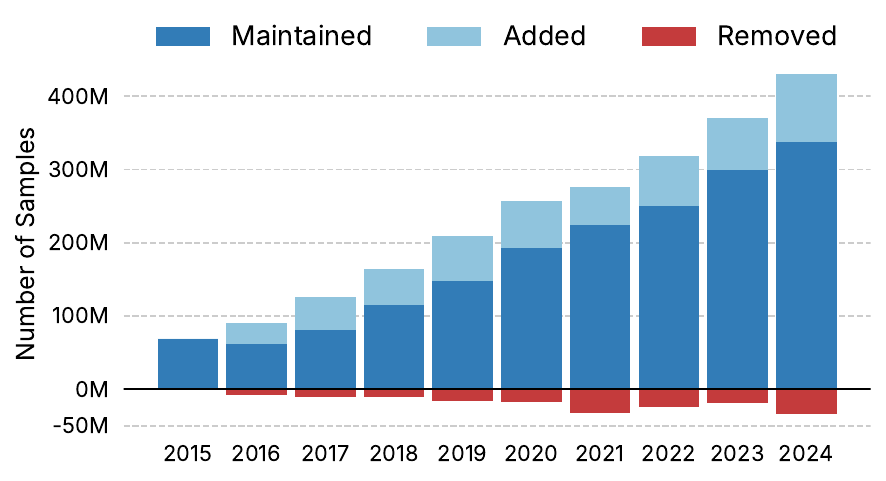}
    \caption{Temporal evolution of UniRef100 from 2015 to 2024. The database shows
	a steady upward trend in total sequences despite millions of sequences being culled
	annually.}
    \label{fig:data_size}
\end{wrapfigure}

The UniProt Knowledgebase~\citep{theuniprotconsortiumUniProtUniversalProtein2025}
(UniProtKB) is the central hub for protein sequence and functional information.
UniProtKB continuously evolves as new sequences are deposited by the community and
existing entries are curated out, either by automated pipelines (e.g., redundant
sequences) or by UniProt curators (e.g., pseudogenes).
In order to track the dynamic
nature of the database, UniProt regularly releases Reference
Clusters~\citep{suzekUniRefClustersComprehensive2015} (UniRef), which are non-redundant
snapshots generated by clustering UniProtKB sequences using the MMseqs2
algorithm~\citep{steineggerMMseqs2EnablesSensitive2017,
kallenbornGPUacceleratedHomologySearch2025}.
Depending on the sequence identity
threshold, i.e., the minimum overlap required for grouping sequences, the resulting
datasets are named as UniRef100, UniRef90, or UniRef50, and only the longest sequence
from each cluster is selected as the representative.
UniRef is the
primary training data for most state-of-the-art protein language
models~\citep{rivesBiologicalStructureFunction2021, linLanguageModelsProtein2022,
fournierProteinLanguageModels2024, nijkampProGen2ExploringBoundaries2023}.

The CoPeP benchmark is composed of 10 consecutive UniRef100 yearly releases from 2015 to
2024, each representing a unique task.
Since a single identifier can correspond to
multiple distinct sequences and a single sequence can be assigned to multiple
identifiers, duplicate entries are removed when both the identifier and sequence are
exactly the same.
Combined, the CoPeP datasets cover 580M unique entries, reflecting
the substantial, nonlinear growth of protein sequence data and shifts in curation
practices (\Cref{fig:data_size}).
The exact releases and size of these snapshots are
provided in \Cref{tab:uniprot}.

As an initial experiment, for each dataset in the benchmark, we compute
the maximum sequence identity for
each sequence in the dataset relative to the sequences
in the
validation set from \Cref{subsec:evaluation} (i.e. we find the closest sequence in the
validation set and compute the sequence identity).
\Cref{fig:cover} (middle and right) shows the evolution of these
sequence identity distributions over years.
We observe a systematic shift toward lower identity values, indicating that newer sequences
are increasingly divergent from the validation set. Furthermore, the year-over-year density shifts
reveal that the initial divergence is driven by substantial shifts in early years,
whereas later transitions shift the distributions back somewhat closer to the validation
set. Appendix~\ref{app:data_evolution} more thoroughly explores the evolution of the data over years.
\begin{figure}
    \centering
    \hfil
	\includegraphics[width=0.34\textwidth]{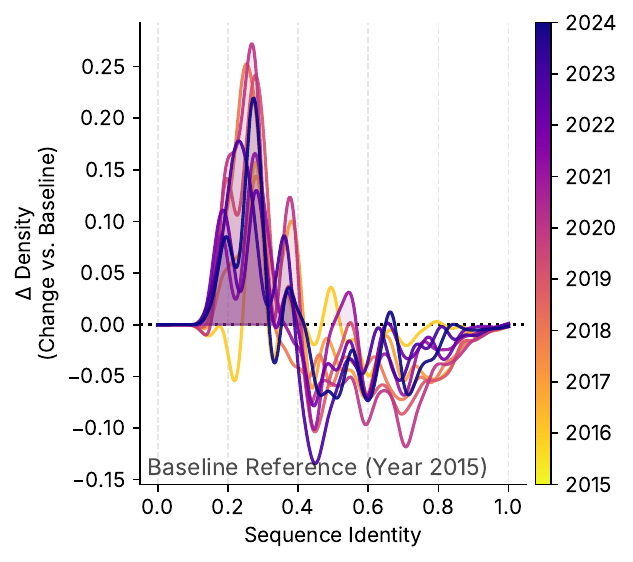}
	\hfil
	\vline
	\hfil
    \includegraphics[width=0.44\textwidth]{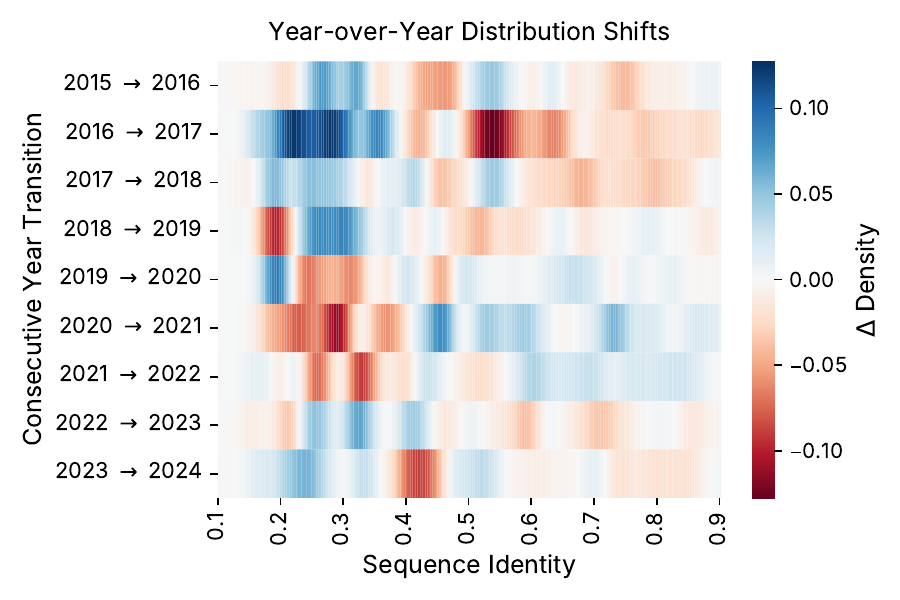}
	\hfil
	\caption{\textbf{Left:} Evolution of sequence identity distributions relative to the validation set. We
        track the sequence identity of the UniRef100 entries in each year compared to their nearest
		neighbor in the validation set. The plot shows the change in sequence identity density
		relative to		
        the 2015 baseline. A systematic shift toward lower identity values (peak at
        $\sim$0.25) indicates that newer sequences are increasingly divergent from the
        validation set. 
        \textbf{Right:} Year-over-year density shifts. The initial
        divergence is driven by substantial shifts in early years, whereas later
        transitions (2020--2024) show smaller, stabilizing fluctuations. }
        \label{fig:seq_identity_evolution}
\end{figure}

\subsection{Streaming Protocol}

In traditional continual learning setups, training proceeds over a sequence ${\di{1},
\dots, \di{n}}$ of $n$ tasks, where each dataset $\di{i} = \{x_j\}_{j=1}^{m_i}$ is drawn
from a task-specific data distribution $x \sim \mathcal{P}_i$.
The challenge typically
arises from distribution shifts between tasks, i.e., $\mathcal{P}_i \neq
\mathcal{P}_{i+1}$, which force the model to balance retaining knowledge of earlier
tasks (stability) with adapting to new tasks (plasticity).

Given the list of $10$ UniRef100 releases $U = [2015, \dots, 2024]$, CoPeP considers
each yearly $U_i$ release as a separate task $\di{i}$.
Compared to traditional continual
learning setups, these different yearly snapshots are all sampled from a shared but
inaccessible underlying distribution of natural proteins $\mathcal{P}^*$.
However, the
individual $\mathcal{P}_i$ distributions exhibit a systematic drift, as the evolution of
protein datasets reflects data growth and shifts in curation practices and research
interests.
Thus, distribution shifts between $\mathcal{P}_i$ and $\mathcal{P}_{i+1}$
still occur.
Additionally, it
is unknown how representative $\di{i}$ is of $\mathcal{P}^*$, with the added challenge
that yearly increments of the dataset do not correlate with improvements of
$\mathcal{P}_i$ w.r.t.
$\mathcal{P}^*$~\citep{fournierProteinLanguageModels2024,spinnerScalingDataSaturation2025}.

Moreover, CoPeP distinguishes itself from traditional continual learning setups by
allowing access to data of past tasks, i.e., the learner may leverage the union of all
observed datasets $\mathcal{U_i} = \bigcup_{j=1}^i\di{j}$.
With this, a learner may
exploit temporal meta-information about the samples.
For example, we can count the
number of consecutive years a protein has persisted in UniRef, i.e., the
\emph{multiplicity} $c(x)$ of a sample, with $c(x)=\sum_{i=1}^k \mathbb{I}_{\di{i}}(x)$.
Such information provides a signal of sequence reliability, distinguishing consistently
preserved proteins from those that appear only transiently.

By structuring the problem this way, CoPeP reflects the practical challenges of
maintaining large-scale models under real-world data evolution, while retaining the core
challenges of continual learning paradigms.

\subsection{Evaluation}
\label{subsec:evaluation}

Unlike traditional continual learning setups, because the underlying distribution
remains constant across tasks, we are not concerned with metrics such as forgetting or
transfer.
Instead, at each evaluation timestep, we measure the performance of the
model on our suite of evaluation tasks at that specific timestep.

\textbf{Validation Set} The validation set curated by
\citet{fournierProteinLanguageModels2024} comprises 10,000 high-quality proteins with
experimental evidence from reference proteomes across all three domains of cellular life
(Bacteria, Archaea, and Eukarya).
The distribution of taxonomy groups is shown in
\Cref{fig:taxonomic_sunburst}.
We perform pairwise deduplication between the validation
set and each yearly release in CoPeP using MMSeqs2 at a sequence identity of 90\%.
While
this threshold is more permissive than the 50\% used by
ESM2~\citep{linLanguageModelsProtein2022}, we argue that they serve fundamentally
different purposes.
Stricter deduplication thresholds such as 50\% are standard for
assessing how well a model generalizes to strictly unseen proteins without overfitting.
In contrast, a 90\% threshold measures how well the model fits the natural distribution
of proteins.
Capturing this distribution is essential for tasks like therapeutic design,
where matching the evolutionary landscape often correlates with stability and
fitness~\citep{fournierProteinLanguageModels2024}.
We track both perplexity and sequence
recovery\footnote{Sequence recovery is defined as the top-1 accuracy of masked token
predictions, where a prediction is correct if the highest probability token matches the
ground truth.} on the validation set, relying on the downstream
benchmarks to evaluate the model's utility and generalization.

\textbf{ProteinGym} ProteinGym~\citep{notin2023proteingym} is a comprehensive fitness
prediction benchmark that evaluates the utility of a model for protein design.
It
comprises millions of mutated sequences from 217 deep mutational scanning (DMS) assays
spanning diverse taxa, including humans, other eukaryotes, prokaryotes, and viruses.
For
each wild-type sequence\footnote{A wild-type sequence is the most common or standard
amino acid sequence for a specific protein found in a natural population, representing
the ``normal'' or unmutated version of that protein.}, the model ranks mutations based
on their predicted fitness.
This ranking is evaluated against experimental ground truth
and clinical annotations using Spearman's rank correlation ($\rho$).
We report the
aggregated performance across all 217 assays.

\textbf{PEER} The benchmark for Protein Sequence
Understanding~\citep{xuPEERComprehensiveMultiTask2022} (PEER) is a multi-task benchmark
designed to evaluate diverse aspects of protein understanding.
Specifically, PEER
comprises 17 tasks covering function, subcellular localization, structural properties,
as well as protein-protein and protein-ligand interactions. We omit the tasks requiring
ligand information as well as the ProteinNet-based
contact prediction task due to its prohibitive computational costs, resulting in a
suite of 14 tasks.

\textbf{DGEB} The Diverse Genomic Embedding
Benchmark~\citep{west-robertsDiverseGenomicEmbedding2024} (DGEB) is another multi-task
benchmark comprising 18 expert-curated datasets spanning sequences from all domains of
life.
DGEB covers a complementary set of biological capabilities, including BiGene
mining, evolutionary distance similarity, clustering, and retrieval.
As our models
operate exclusively on amino acid sequences, we restrict our evaluation to the 16 tasks
utilizing this modality.

\subsection{Base Experimental Setup}
\label{subsec:base}

We adopt the bi-directional AMPLIFY 120M~\citep{fournierProteinLanguageModels2024} as
our base pLM for its computational efficiency.
Following the experimental setup of
\citet{fournierProteinLanguageModels2024}, we train for 100k steps per task using
AdamW~\citep{loshchilovDecoupledWeightDecay2018} with a weight decay of $10^{-2}$, an
effective batch size of $4,096$, and a maximum sequence length of $512$.
In order to
mitigate the difficulty associated with re-warming the learning rate in continual
learning~\citep{guptaContinualPreTrainingLarge2023}, we replace the cosine decay with
Warmup-Stable-Decay~\citep{huMiniCPMUnveilingPotential2024,
liModelMergingPretraining2025} (WSD).
As shown in \Cref{fig:wsd_v_cos}, this
modification does not increase the validation perplexity compared to the cosine baseline
(5.44 for cosine vs 5.43 for WSD).
Specifically, we linearly warm up the learning rate
from $0$ to $5 \times 10^{-4}$ over the first 10k steps, keep it constant until 90k
steps, after which it is decayed linearly from $5 \times 10^{-4}$ to $0$.
When
transitioning tasks, we discard the previous decay phase and reset the model to the
pre-decay checkpoint.
Consequently, the cumulative effective training steps at the start
of the $n$-th task is $(n-1) \times 90k$.
For instance, training on the sixth task
begins from a checkpoint with 450k effective gradient steps.

\subsection{Motivation for Temporal Meta-Information}
\label{sec:filter}
To highlight the value of temporal metadata and motivate our continual learning
benchmark, we investigate the hypothesis that temporal persistence is a proxy for the
likelihood of entries being functional proteins.
Specifically, we posit that sequences
retained in UniProtKB over releases provide a less noisy, more robust learning signal.
For each pair of years within our dataset, we train models
on only the sequences present in both releases.
All models are trained for
100k steps following the protocol outlined in \Cref{subsec:base}.

\begin{figure}
    \centering
    \hfil
    \includegraphics[width=.45\linewidth]{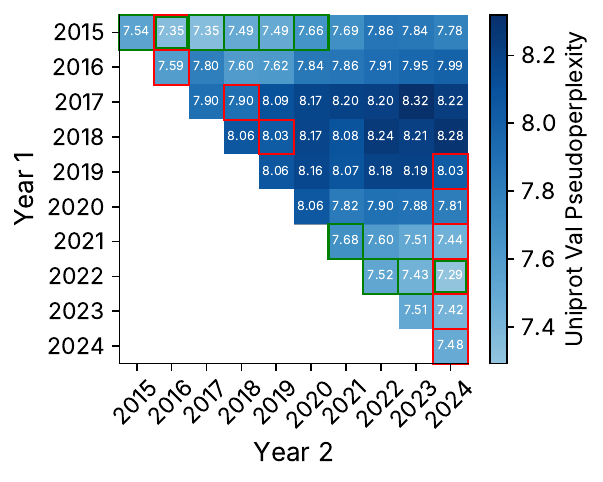}
    \hfil
    \includegraphics[width=.45\linewidth]{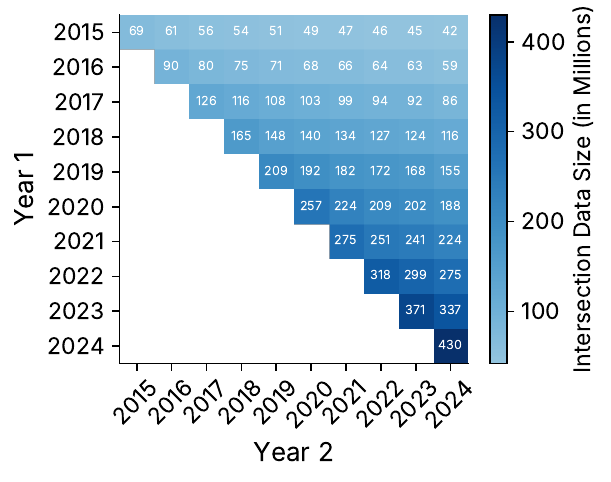}
    \hfil
    \caption{Impact of temporal data filtering.
Models are trained exclusively on the
    intersection of sequences retained between two UniRef100 releases.
Left shows the validation perplexity for each model and right shows the corresponding
dataset sizes. 
The best performance in each row is boxed in 
red, and the best performance in each column is boxed in green.
    Most years show improved validation perplexity on at least one intersection
    dataset despite a substantial reduction in dataset size. 
    The best performance by any model is achieved by a model
    trained on the intersection of 2022 and 2024, despite only having 64\% of the
    data of the larger 2024 release.
    } 
	\label{fig:filter}
\end{figure}

\Cref{fig:filter} reports the validation perplexity for each model, with the
diagonal representing baselines trained on unfiltered annual releases.
The effect of filtering is not always better than the full year baseline, 
likely due to the competing effects of reduced dataset size and improved data quality.
Despite this, for 7/9 columns and 8/9 rows, we observe that at least one intersection
outperforms the corresponding full year baseline. 
The best performance is achieved by the intersection of
2022 and 2024 data.
In fact, the model achieves a 2.5\% improvement in perplexity over the full 2024
release, despite only having access to 64\% of the data,
supporting the utility of temporal
persistence as a signal for curating pretraining corpora.
The heatmap gives us two perspectives on using temporal information for data curation.
Looking up columns tells us the optimal way of filtering our current dataset using a
past year, while looking across rows gives us a signal that there is potential
value in continual learning and looking at the temporal evolution of the data.

\section{Methods}
\label{sec:methods}
We now perform a large-scale study
of 6 different methods to continually pretrain the AMPLIFY-120M model.
We focus on a set of representative methods spanning across 3 groups:
standard continual learning, plasticity-preserving, and unlearning.
We compare these approaches to a \textbf{Joint} model trained on the
cumulative data from 2015 to 2024 and 
to two types of single year baselines (i.e. models trained 
only on one year's data): \textbf{Incremental} models are trained for 100k
steps each year, and \textbf{Matched} models are trained for the number of steps equal to
the cumulative steps of the continual learning methods up to that year (e.g. 370k steps
for the fourth year in the benchmark). 

\subsection{Continual Learning}

\textbf{Sequential Training} This approach serves as the naive baseline, in which the pLM
is trained sequentially on each yearly release for 100k steps without any additional
interventions or regularization.
Consequently, this method introduces no additional hyperparameters beyond the base
experimental setup.

\textbf{Temporal Replay} Standard Experience
Replay~\citep{rolnickExperienceReplayContinual2019a, abbesRevisitingReplayGradient2025}
mitigates catastrophic forgetting by rehearsing a small, fixed-size subset of data from
prior tasks alongside the current one.
Given the availability of historical releases and based on our findings in
\Cref{sec:filter}, we modify this approach to use an unbounded replay buffer with a
temporal importance sampling strategy.
Formally, let $S = \{\di{1}, \di{2}, \ldots, \di{t-1}\}$ be the sequence of datasets up
until the current task, and let $U = \bigcup_{i=1}^{t-1} \di{i}$ be their union.
For any example $x \in U$, let its multiplicity be $c(x) = \sum_{i=1}^k
\mathbb{1}_{\di{i}}(x)$, where $\mathbb{1}_{\di{i}}(x)$ is the indicator function.
During training, we sample replay examples from $U$ with probability proportional to
their multiplicity:
\begin{equation}
P(x) = \frac{c(x)}{\sum_{y \in U} c(y)} = \frac{\sum_{i=1}^k \mathbb{1}_{D_i}(x)}{\sum_{i=1}^k |D_i|}.
\end{equation}
The optimization objective is a weighted combination of the loss on the new data and the
historical replay:
\begin{equation}
	\mathcal{L} = (1-\lambda_{replay})\mathcal{L}_{ce}(b_{i}) + \lambda_{replay}\mathcal{L}_{ce}(b_{replay})
\end{equation}
where $b_i$ is a batch of sequences introduced in the current task $i$, $b_{replay}$ is
a batch sampled from previous tasks according to $P(x)$, and $\lambda_{replay}$ is a
hyperparameter controlling the importance of replay.

\subsection{Plasticity}

Loss of plasticity refers to the phenomenon in continual learning in which a model's ability
to adapt to changes in data distributions degrades as training progresses.
The plasticity-preserving methods considered in our experiments are agnostic to prior data
distributions and do not rely on any additional data.

\textbf{Shrink and Perturb} Shrink and
Perturb~\citep{ashWarmStartingNeuralNetwork2020} mitigates the loss of plasticity by
periodically scaling down network weights and injecting noise.
We apply this intervention at the start of each task by updating the weights as
$\theta_t = \lambda_{shrink}\theta_{t-1} + \lambda_{noise}p$, where $p$ represents
random noise sampled from the network's original initialization distribution.

\textbf{Hare and Tortoise} Hare and Tortoise~\citep{leeSlowSteadyWins2024} maintains
two parallel sets of network weights: fast and slow.
The slow weights are an exponential moving average of the fast weights, updated at every
step as $\theta_{slow} = \lambda_{ht\_mom}\theta_{slow} +
(1-\lambda_{ht\_mom})\theta_{fast}$.
Periodically, the fast weights are reset to the slow weights according to
$\lambda_{reset\_freq}$.

\subsection{Unlearning}
Unlearning aims to actively erase knowledge about specific samples from the network.
In our experiments, the forget set for task $t$ is denoted as $\ft{}$ and defined as the
set of sequences present in task $t-1$ but not in task $t$.
At each step, we sample one batch from the current task $b_i \sim \dt{}$ and one from
the forget set $b_{forget} \sim \ft{}$.

\paragraph{Gradient Ascent} Gradient Ascent~\citep{golatkarEternalSunshineSpotless2020}
takes a dual-objective approach to unlearning: concurrently maximizing the loss on the
forget set to erase knowledge while minimizing the loss on the current task to acquire
new information and prevent divergence.
This strategy is implemented by minimizing the objective $\mathcal{L}_{ce}(b_{i}) -
\lambda_{asc}\mathcal{L}_{ce}(b_{forget})$, where $\mathcal{L}_{ce}$ is the
cross-entropy loss and $\lambda_{asc}$ controls the strength of the unlearning.

\textbf{Random Labels} Random Labeling~\citep{golatkarEternalSunshineSpotless2020}
achieves unlearning by corrupting the model’s knowledge of the forget set.
Specifically, it overwrites learned correlations by replacing ground-truth targets with
random tokens sampled uniformly from the vocabulary, thereby training the model to
predict noise.
The cross-entropy loss is minimized against these noisy targets, and this unlearning
objective is added to the standard task loss, weighted by $\lambda_{rand}$.

\section{Results}

\subsection{UniProt Validation Set}

\begin{figure}
	\centering
	\includegraphics[width=\linewidth]{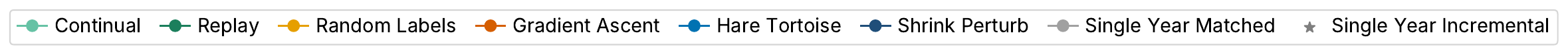}
	\\
	\includegraphics[width=.48\linewidth]{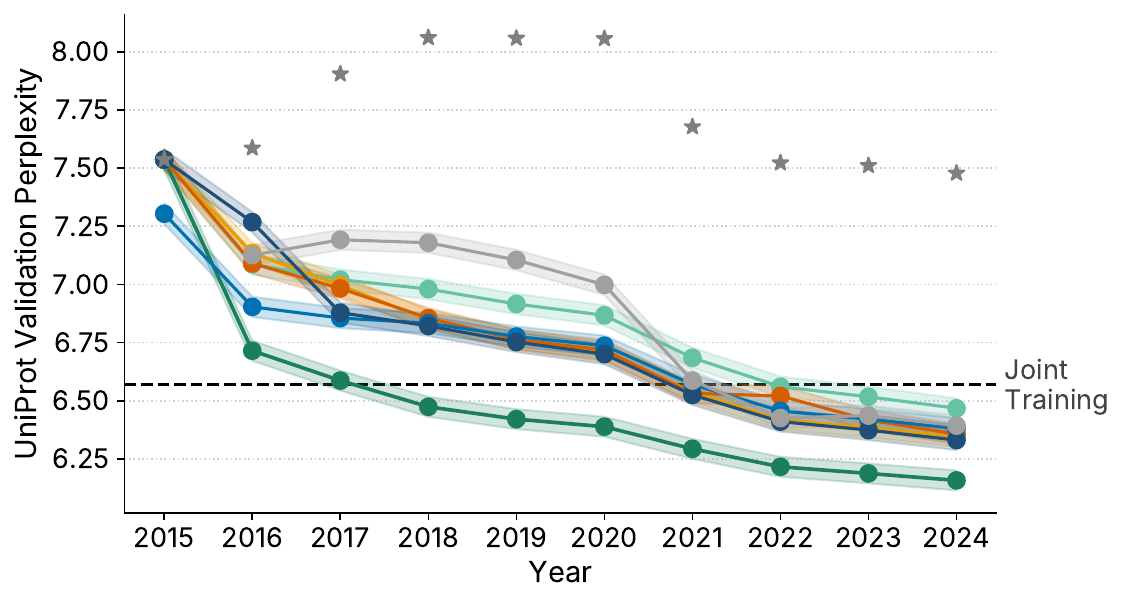}
	\includegraphics[width=.48\linewidth]{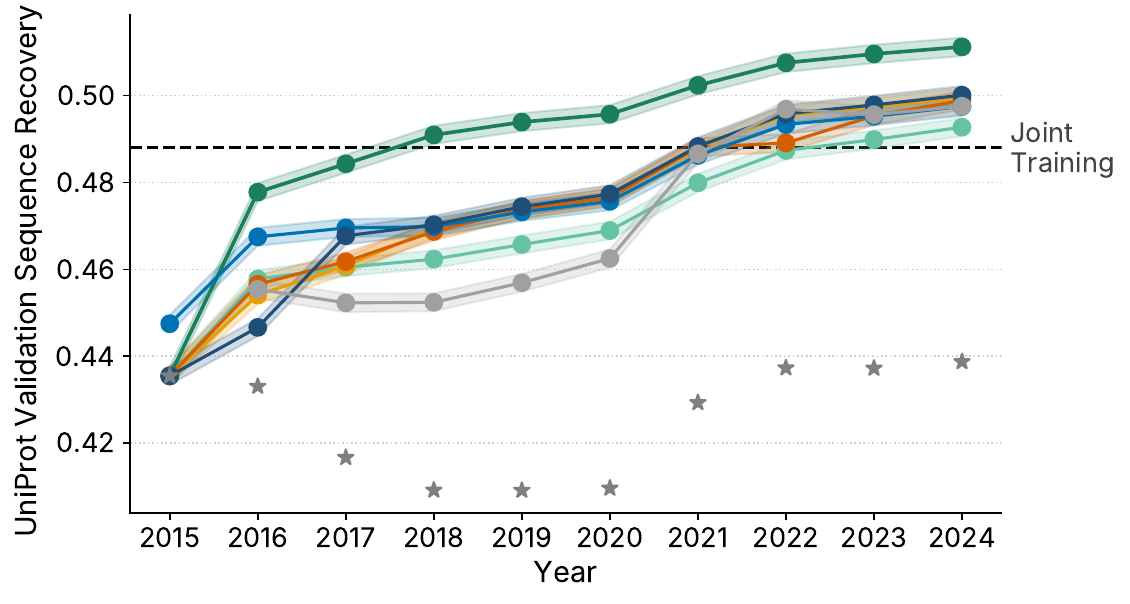}
	\caption{Perplexity (left) and Sequence Recovery (right) on the UniProt validation
	set with shading representing standard error.
	Continual learning baselines generally improve over time, and outperform jointly
	training on all data.
	Temporal Replay achieves the best performance outperforming other continual methods
	and single year matched training for all years. Every continual baseline
	outperforms	the naive continual learning baseline, validating many of them for the
	first time at scale.
}
	\label{fig:perplexity}
\end{figure}

The perplexity and sequence recovery on the UniProt validation set shown in
\Cref{fig:perplexity} show that the performance of the continual learning baselines
generally improves over time.
While this might appear minor, it validates taking a continual approach to the
problem and shows that continual training does not saturate the network or
prime the network so heavily that it cannot learn from future data.
Every continual method outperforms the Single Year Incremental baselines, the joint
training baseline, and, in many cases, even the Single Year Matched baselines.
The improvement over incremental baselines establishes that in a compute
constrained setting, restarting from a continual checkpoint is more effective than
starting from scratch. The fact that several methods outperform the Single Year
Matched baselines shows that even in a compute rich setting, continual learning
can provide benefits over standard training practices.
The improvement over the joint training baseline, normally a very strong
baseline in continual learning settings, points to something unique about the sequential
protein data setup. Because the joint model learns from sequences that were
subsequently removed from the database, often due to being redundant or pseudogenes, it
might in fact be suboptimal for learning a good protein distribution.
Since the validation set is curated to contain strictly high-quality, experimentally
verified sequences, the continual models, which filter out these removed entries, learn
a distribution that is better aligned with valid protein data. 

Finally, comparing the methods amongst each other, we see that every method offers
better performance compared to the naive continual baseline and (other than the temporal
replay baseline)
relatively similar performance to each other. This is highly encouraging, as
none of these methods were developed for this specific setup.
Hare and Tortoise and Shrink and Perturb are both plasticity preserving methods, but to
our knowledge have never been applied to a model or training scale of this size.
Gradient Ascent and Random Labels have been used with LLMs, but generally on more
limited forget sets and not as a part of
continual pretraining. 
The relative success of these diverse strategies suggests that the core paradigms of
continual learning (forgetting, plasticity, and unlearning) offer valuable contributions
to protein modeling.

\subsection{ProteinGym}

\begin{figure}
	\centering
	\includegraphics[width=.5\textwidth]{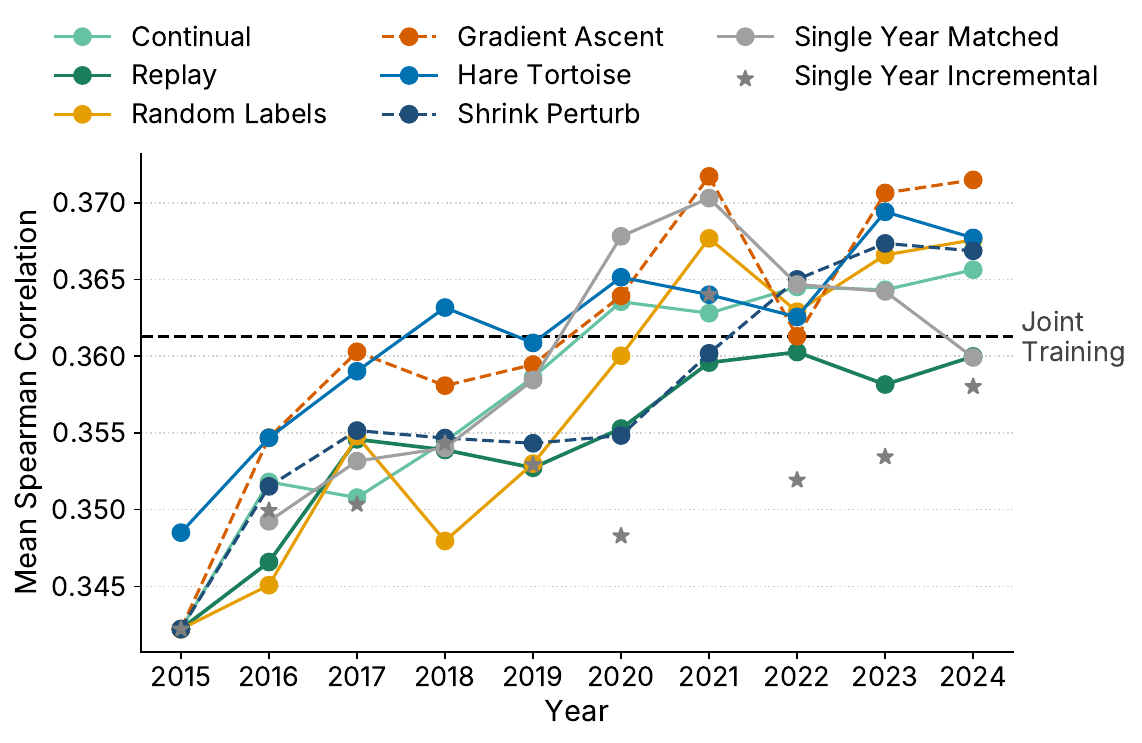}
	\caption{Spearman correlation on ProteinGym.
	5 of the 6 continual learning methods outperform joint training, with Gradient
	Ascent and Hare Tortoise achieving the best scores. The best single year performance
	(2021 matched) is also outperformed by Gradient Ascent.
}
	\label{fig:proteingym}
\end{figure}

In contrast to the trends observed on the UniProt validation set, the results on
ProteinGym shown in \Cref{fig:proteingym} reveal a more nuanced picture.
First, nearly every continual method outperforms Joint Training, with Gradient Ascent
and Hare Tortoise consistently emerging as the top-performing methods.
Moreover, four continual strategies outperform the naive sequential training baseline,
reaffirming the benefit of specialized continual learning methods.

Temporal Replay, however, significantly lags behind on ProteinGym despite achieving the
best performance on the UniProt validation set.
We explore potential reasons for this discrepancy in \Cref{subsec:tradeoffs}.
Specifically, the UniProt validation set is skewed towards highly studied proteomes,
sequences that persist over time and are therefore prioritized by our replay mechanism.
In contrast, ProteinGym evaluates the model's ability to predict fitness effects for
specific, often heterogeneous mutations.
Therefore, methods that emphasize historical sequences may not perform well on
downstream tasks such as ProteinGym whose distribution does not closely align.

\subsection{Multi-Task Protein Understanding Benchmarks}
\begin{figure}
	\centering
    \includegraphics[width=.3\linewidth]{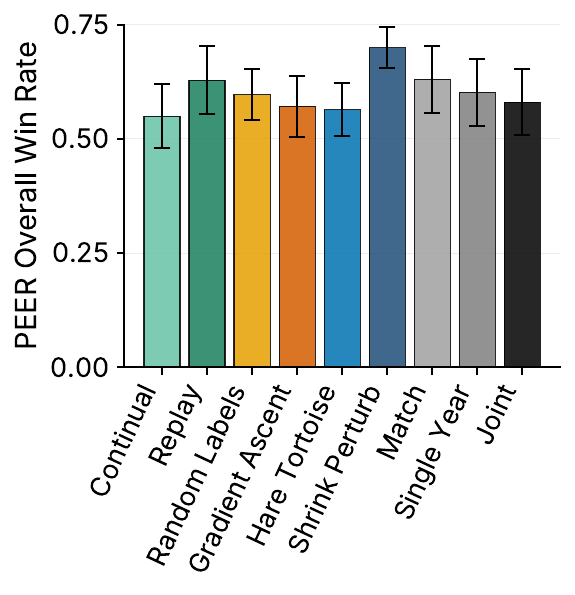}
	\hfil
    \includegraphics[width=.3\linewidth]{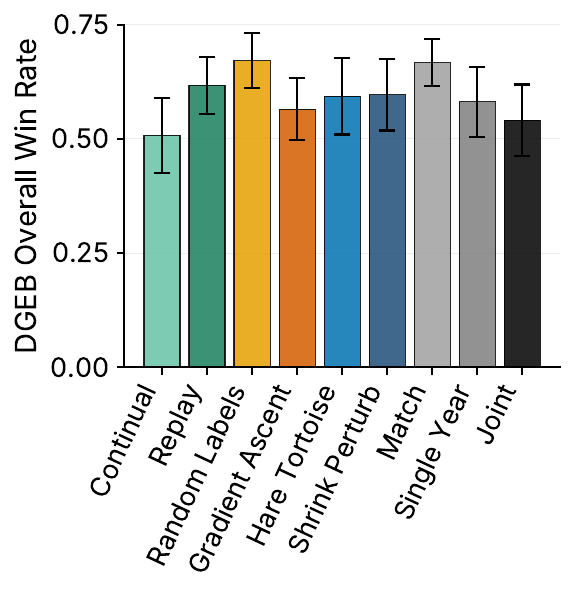}
	\caption{Win rates on multi-task protein understanding benchmarks for the
	best checkpoint for each method.
	\textbf{Left}: On PEER, Shrink and Perturb achieves the highest
	win rate, with Replay and Single Year Match in second place.
	\textbf{Right}: On DGEB, Random Labels achieves the highest win rate.
	Across both benchmarks, specialized continual learning methods consistently match or
	outperform the naive continual and single year baselines.}
    \label{fig:peer_dgeb}
\end{figure}

For the multi-task protein understanding benchmarks PEER and DGEB, each task in the
benchmark has a different metric, making it challenging to aggregate results across
tasks. Thus, we instead compute a win rate, defined as the proportion of times a
model outperforms the other models across all tasks and years.
This win rate is computed against every other checkpoint across all methods, years, and
tasks, making it a relatively stable metric.
In \Cref{fig:peer_dgeb}, we present the checkpoint with the highest win rate for each
method.
Fine-grained results are provided in \Cref{appendix:PEER} and \labelcref{appendix:DGEB}.

Across both benchmarks, specialized continual learning strategies clearly outperform the
naive continual baseline, especially on DGEB where the naive approach performs poorly.
Shrink and Perturb achieves the highest win rate on PEER, while Random Labels leads on
DGEB.
Notably, Temporal Replay demonstrates a strong and consistent performance on both
benchmarks.

Regarding the Single Year baselines, performance for the Match baseline is high across
both benchmarks, only slightly trailing the best continual method. Performance for the
Incremental baseline is approximately on par with the majority of the continual methods,
suggesting that simply training for longer may not be sufficient to achieve strong
performance on these tasks.

Finally, while the results on DGEB and PEER generally align with one another, they
diverge significantly from the trends observed in ProteinGym.
This disconnect suggests that the optimal data characteristics required for general
protein understanding differ distinctively from those required for zero-shot fitness
prediction.

\section{Discussion}
\label{subsec:tradeoffs}

Drug discovery is a complex, multi-stage process that requires a diverse set of
capabilities from the pLM, and designing a single continual learning method that
performs well across all tasks is challenging. Interestingly, we observe that different 
methods excel on different benchmarks, highlighting the diverse requirements of these
tasks. Replay did well on the UniProt validation set, while Gradient Ascent and Hare and
Tortoise
performed best on ProteinGym. Shrink and Perturb and Random Labels achieved the highest
win rates on PEER and DGEB, respectively. 
In this section, we discuss the specific trade-offs of the benchmarks used in our study.

The UniProt validation set is designed to reflect the natural distribution of protein
sequences, assuming that evolutionary fitness provides a valuable signal.
Because this set comprises high-quality sequences with experimental evidence, it is
skewed toward well-studied proteomes that were sequenced early and have persisted in the
UniProt Knowledgebase.
This bias likely explains the strong performance of Temporal Replay, which prioritizes
such persistent sequences.

In contrast, ProteinGym evaluates the sensitivity of pLMs to local changes in the
fitness landscape by ranking specific mutations.
Unlike the UniProt validation set, we did not perform strict deduplication between the
training corpora and the wild-type sequences found in ProteinGym, as is common.
Consequently, methods that allow memorizing specific sequences or overfitting the
neighborhood of wild-type proteins present in UniRef100 may
have an advantage.

Finally, PEER and DGEB measure the transferability of learned representations by
fine-tuning them on a broad set of downstream tasks.
These benchmarks require that pLMs encode high-level biological features into their
embeddings, rather than sequence likelihoods or local fitness landscapes. PEER is
designed for more general protein understanding tasks, while DGEB focuses on 
diverse and less well-studied genomic proteins.
Thus, methods that encourage broader generalization, such as Shrink and Perturb
do well on PEER, and methods that perhaps push the model away from overfitting to
well-studied sequences, such as Random Labels, excel on DGEB.

\section{Conclusion}

We introduce CoPeP, a large-scale benchmark that leverages the temporal evolution of the
UniProt Knowledgebase to evaluate continual pretraining strategies on a complex,
real-world problem.
Our extensive evaluation reveals that specialized continual learning methods
consistently outperform naive sequential baselines.
Temporal Replay, in particular, outperforms all continual and non-continual baselines,
by leveraging sequence persistence.
However, we observed a critical trade-off depending on the downstream application: while
replay-based strategies excel at modeling natural protein distributions,
plasticity-preserving and unlearning approaches like Hare and Tortoise and Gradient
Ascent performed better for zero-shot fitness prediction on ProteinGym,
while Shrink and Perturb and Random Labels offered better transfer
learning on PEER and DGEB.
Future work could explore how to combine these orthogonal methods to develop more
effective approaches.
Ultimately, CoPeP demonstrates that temporal metadata can be leveraged to maintain
state-of-the-art performance without expensive retraining, paving the way for more
sustainable and accessible drug discovery research.
We hope this benchmark will accelerate progress in protein language models.

\section{Reproducibility Statement}

The details of our model training and hyperparameter selection are provided in
\Cref{subsec:base,app:lr,app:hparam}. The details of our dataset curation are provided in
\Cref{subsec:dataset,subsec:evaluation,app:data}. 
We have also released the code\footnote{\url{https://github.com/chandar-lab/CoPeP}} as well
as the checkpoints and datasets
\footnote{\url{https://huggingface.co/collections/chandar-lab/CoPeP}} used to conduct
all of our experiments. 

\section{Acknowledgements}
This research was enabled in part by compute resources provided by Calcul Québec
(calculquebec.ca), the Digital Research Alliance of Canada (alliancecan.ca), and Mila
(mila.quebec). 
SC is supported by the
Canada CIFAR AI Chair, the Canada Research Chair in Lifelong Machine Learning, and the
NSERC Discovery Grant.
We would also like to thank Genentech for their funding and support for this project.
Finally, we would like to thank David Heurtel-Depeiges and Lola Le Breton for their help
with the initial data curation and experimental setup.

\bibliography{bibliography}
\bibliographystyle{iclr2026_conference}
\clearpage
\appendix
\section{Evolution of Data}
\label{app:data_evolution}
\subsection{Embedding Visualization}
\label{sec:embedding_viz}
In this experiment, we analyze how protein sequence datasets evolve over years by visualizing their structure in embedding space. We extract the embeddings from the last layer of AMPLIFY 350M (trained with 1 million steps) \citep{fournierProteinLanguageModels2024} and apply UMAP~\citep{umap} to project high-dimensional protein embeddings into two dimensions. This enables us to observe broad patterns in the data and how they change across consecutive UniRef100 releases.

\begin{figure}[!htb]
	\centering
	\includegraphics[width=0.4\textwidth]{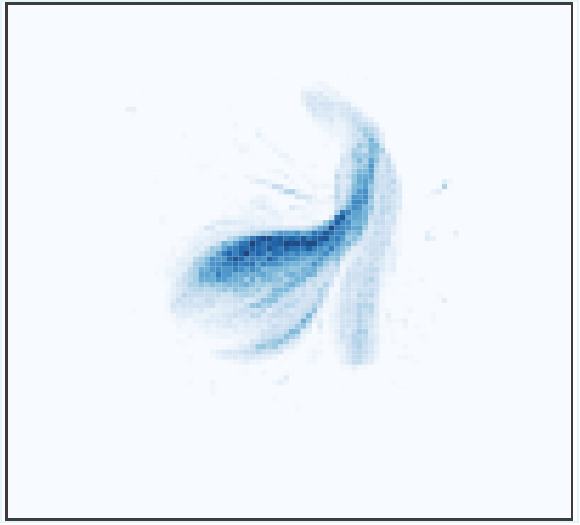} \\
	\caption{UMAP projection of protein embeddings from all UniRef100 releases (AMPLIFY-350M representations). The plot shows a stable global structure with a dense core and branches, indicating natural groupings of proteins.}
	\label{fig:emb_viz_full}
\end{figure}

\Cref{fig:emb_viz_full} shows the embedding of the full dataset, i.e., sequences from all years. The plot reveals a global structure with a dense central region and branches, suggesting natural groupings of proteins. Differences in density highlight areas where certain types of sequences are more common.

\begin{figure*}[htb!]
	\centering
	\includegraphics[width=\textwidth]{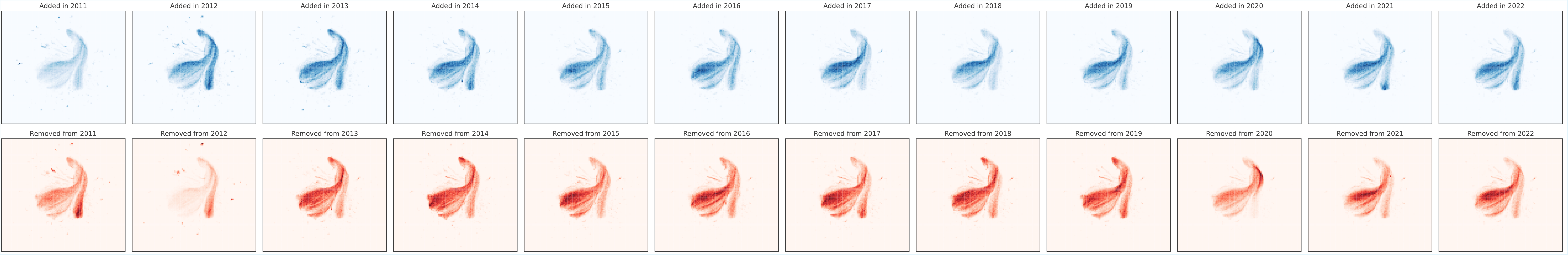}
	\caption{Yearly dynamics of UniRef100 embeddings. Top row (blue): proteins added in each year; bottom row (red): proteins removed. While the global organization of protein embeddings is stable, the local shifts such as density increases and cluster expansion indicate yearly shifts in the underlying distribution.}
	\label{fig:emb_viz}
\end{figure*}

Each UniRef100 release both adds and removes sequences, reflecting the expansion of available protein sequences and ongoing curation. To illustrate these dynamics, \Cref{fig:emb_viz} compares additions (blue, top row) with removals (red, bottom row) per year. Overall, while the global structure of protein embeddings is stable, \Cref{fig:emb_viz} indicates local shifts such as density increases and cluster expansion. This underscores why continual learning is critical for protein language models. Instead of treating each release as an isolated dataset, continual methods can exploit temporal information to adapt to new proteins while retaining prior knowledge.

\subsection{Model Embedding Shifts}

\begin{figure*}[htb!]
	\centering
	\includegraphics[width=\textwidth]{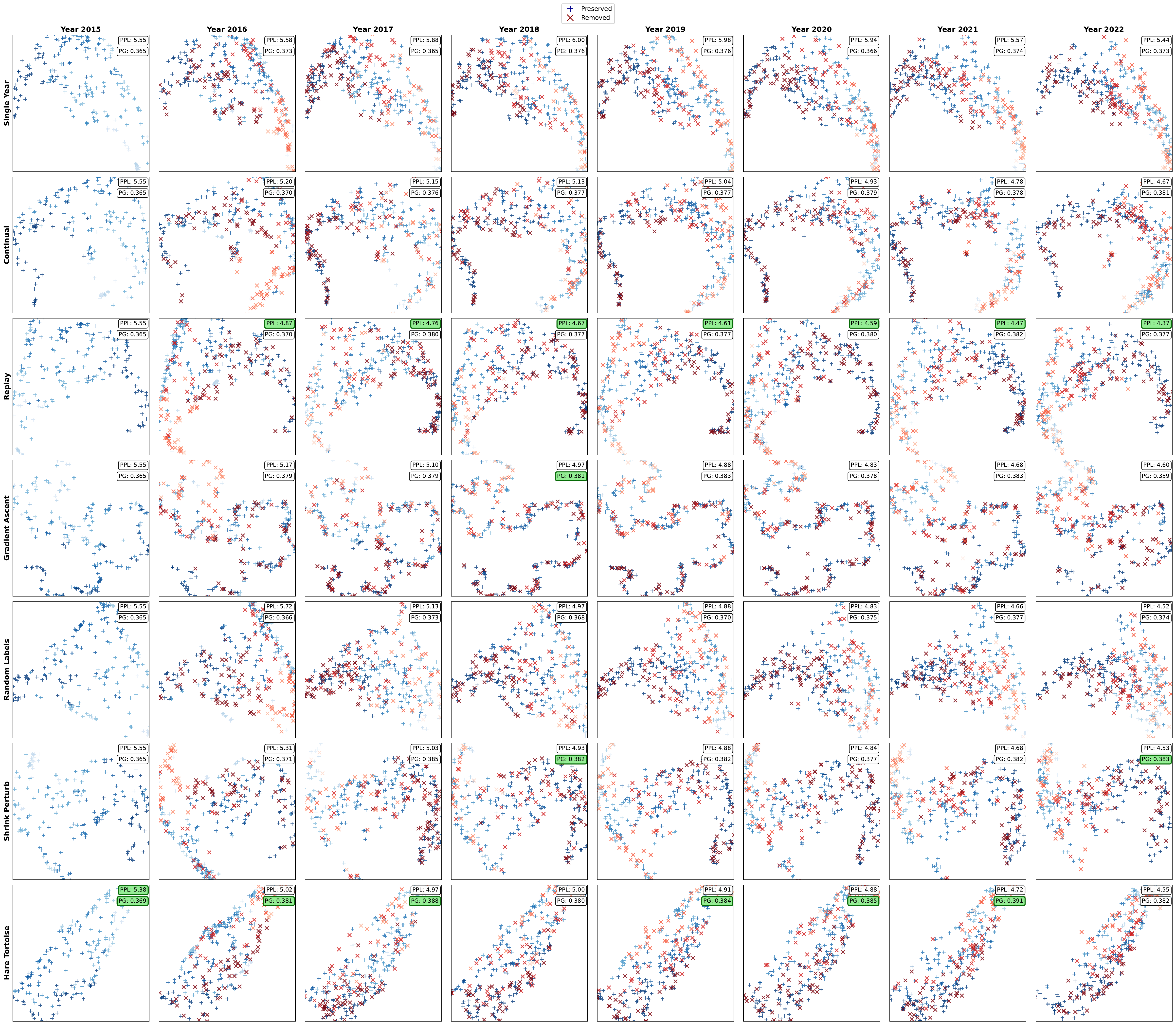}
	\caption{UMAP visualization of protein embeddings across continual learning methods
		$(2015–2022)$ with performance metrics (PPL: validation perplexity; PG: ProteinGym
		mean Spearman correlation). Color shades are based on Single Year clusters. Darker
		shades indicate denser regions. The same color shades are used in other rows/methods to
		indicate those centroid/representative protein locations.
	}
	\label{fig:umap_methods}
\end{figure*}

In \Cref{fig:umap_methods}, we visualize how different continual learning methods
structure their protein embeddings. For each method, we select the final checkpoint and
extract embeddings for a representative subset of proteins that were added and removed
across all years. We then apply UMAP to project these embeddings into two dimensions.
The color shading indicates protein density within the Single Year clusters, with darker
shades indicating denser regions. Interestingly, most methods maintain a similar global
structure except for Gradient Ascent, which shows a more rope-like structure.

\subsection{Data Statistics}
\label{sec:data-statistic}

\begin{figure*}[htb!]
	\centering
	\begin{subfigure}[t]{.32\textwidth}
		\centering
		\includegraphics[width=\linewidth]{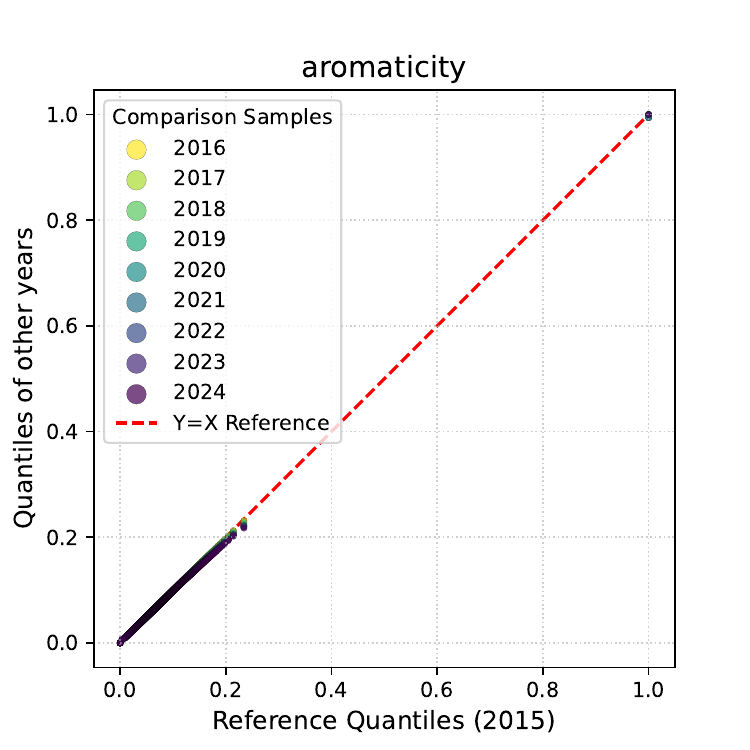}
		\caption{Aromaticity}
		\label{fig:qq_aromaticity}
	\end{subfigure}
	\hfill
	\begin{subfigure}[t]{.32\textwidth}
		\centering
		\includegraphics[width=\linewidth]{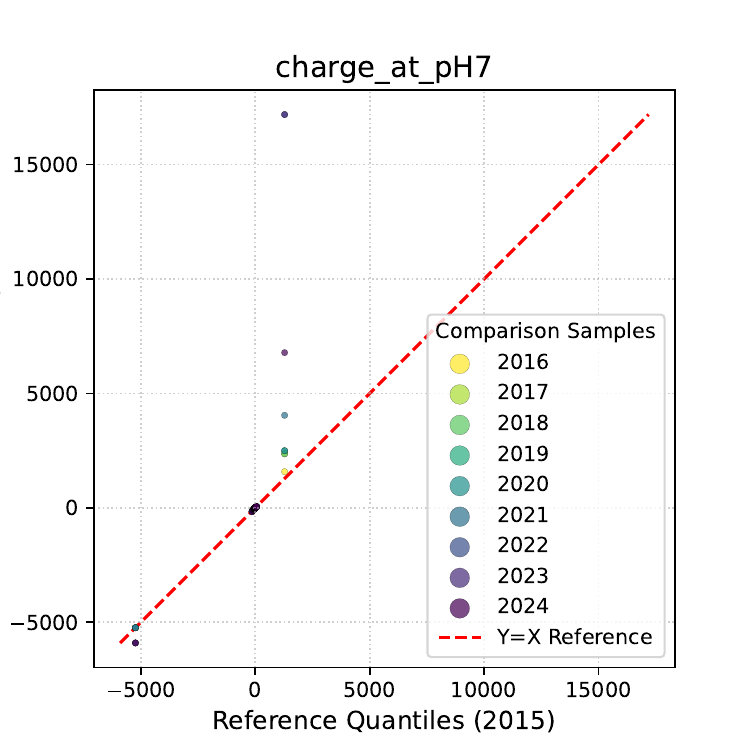}
		\caption{Charge at pH 7}
		\label{fig:qq_charge}
	\end{subfigure}
	\hfill
	\begin{subfigure}[t]{.32\textwidth}
		\centering
		\includegraphics[width=\linewidth]{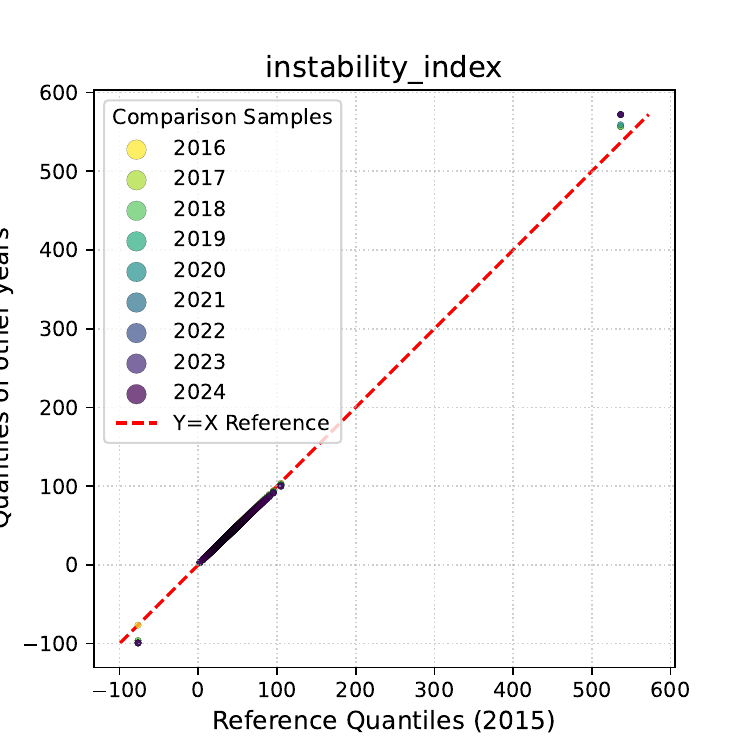}
		\caption{Instability Index}
		\label{fig:qq_instability}
	\end{subfigure}
	\begin{subfigure}[t]{.32\textwidth}
		\centering
		\includegraphics[width=\linewidth]{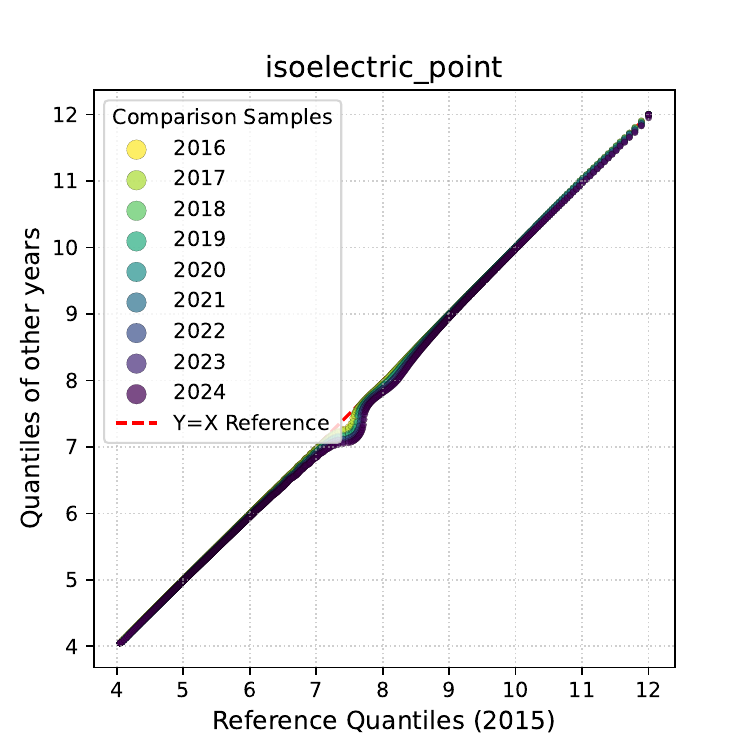}
		\caption{Isoelectric Point}
		\label{fig:qq_isoelectric_point}
	\end{subfigure}
	\hfill
	\begin{subfigure}[t]{.32\textwidth}
		\centering
		\includegraphics[width=\linewidth]{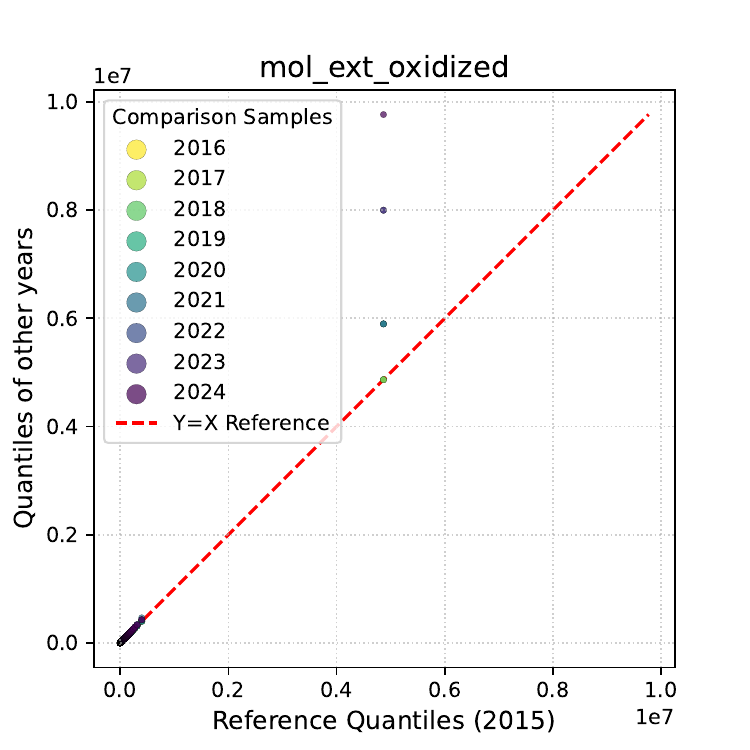}
		\caption{Molar Extinction Coefficient (oxidized)}
		\label{fig:qq_mol_ext_oxidized}
	\end{subfigure}
	\hfill
	\begin{subfigure}[t]{.32\textwidth}
		\centering
		\includegraphics[width=\linewidth]{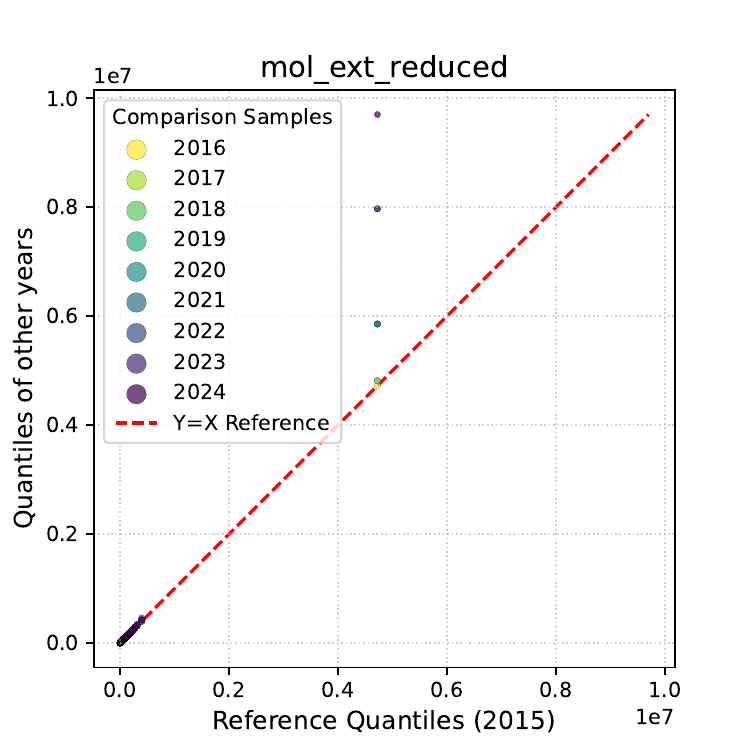}
		\caption{Molar Extinction Coefficient (reduced)}
		\label{fig:mol_ext_reduced}
	\end{subfigure}
	\hfill
	\begin{subfigure}[t]{.32\textwidth}
		\centering
		\includegraphics[width=\linewidth]{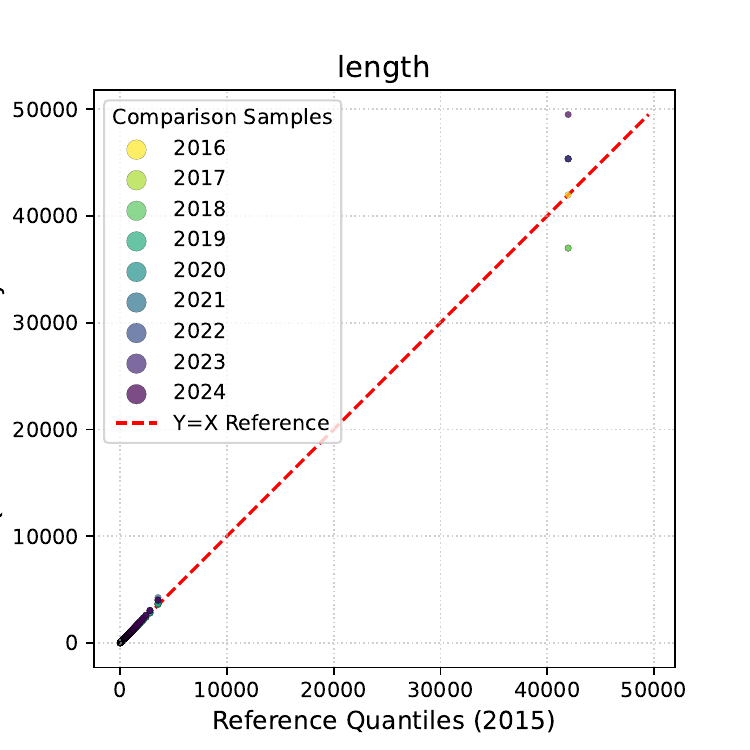}
		\caption{Protein Length}
		\label{fig:qq_length}
	\end{subfigure}
	\begin{subfigure}[t]{.32\textwidth}
		\centering
		\includegraphics[width=\linewidth]{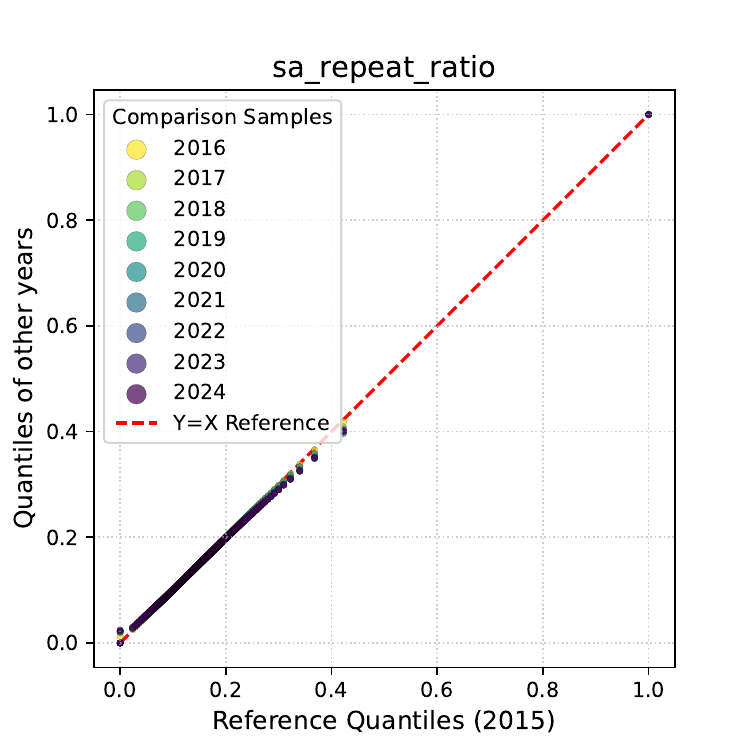}
		\caption{Longest Repeat Ratio}
		\label{fig:qq_repeat_ratio}
	\end{subfigure}
	\hfill
	\caption{Two sample QQ-plots of various protein sequence statistics across different
	years. Each line represents a comparison between a year and the reference year 2015.}
\end{figure*}

In this section, we analyze how various protein sequence simple biochemical properties
evolve over years. We compute several statistics for each protein sequence in the
UniRef100 releases from 2015 to 2024 using the Biopython
library~\citep{cockBiopythonFreelyAvailable2009}. Specifically, we create two-sample
QQ-plots comparing the distribution of each statistic between the reference year 2015
and each yearly release. The statistics we analyze include:
\begin{multicols}{3}
	\begin{itemize}
		\item Aromaticity
		\item Charge at pH 7
		\item Instability Index
		\item Isoelectric Point
		\item Molar Extinction Coefficient (oxidized and reduced)
		\item Protein Length
		\item Longest Repeat Ratio
	\end{itemize}
\end{multicols}
Overall, we see that most statistics show relatively stable distributions across years,
for at least some of the statistics (molar extinction coefficient, protein length,
charge at pH 7) there are outliers at the extremes of the distributions for several
years. Furthermore, there is a noticeable shift in the distribution of the isoelectric
point over the years, with later years showing a higher density of proteins with
isoelectric points around 7-8. 
Despite these biophysical properties of the sequences
(calculated using fairly simple, surface level statistics) remaining relatively stable,
we can see from the embedding visualizations in \Cref{sec:embedding_viz} and the 
sequence identity analysis in \Cref{fig:seq_identity_evolution} that the distribution of
sequences is shifting significantly over years.

\section{UniProt Validation Results}
\subsection{Additional Filter Experiment Results}
\begin{figure*}
	\centering
	\includegraphics[width=.48\textwidth]{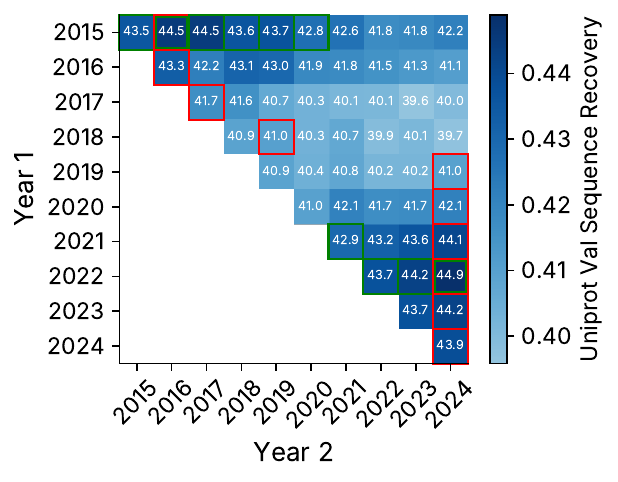}
	\caption{Accuracy on the UniProt validation set when training on
	different filtered datasets.
	The best performance in each row is boxed in red, and the best performance in each
	column is boxed in green.
	Note, the best performance overall ($2022 \cap 2024$) corresponds to training on an
	intersection that is 64\% the size of the largest dataset (2024 no filter).
	}
	\label{fig:data_intersection_size}
\end{figure*}

In \Cref{fig:data_intersection_size}, we present additional results from the dataset
filtering experiments described in \Cref{sec:filter}. Specifically, we show sequence
recovery on the UniProt validation set when training on different filtered
datasets.

\subsection{UniProt Dataset Composition}
\label{app:uniprot_val}
In this section, we provide additional information on the makeup of the training and
validation set used in throughout the paper.

\begin{figure*}
	\centering
	\includegraphics[width=\linewidth]{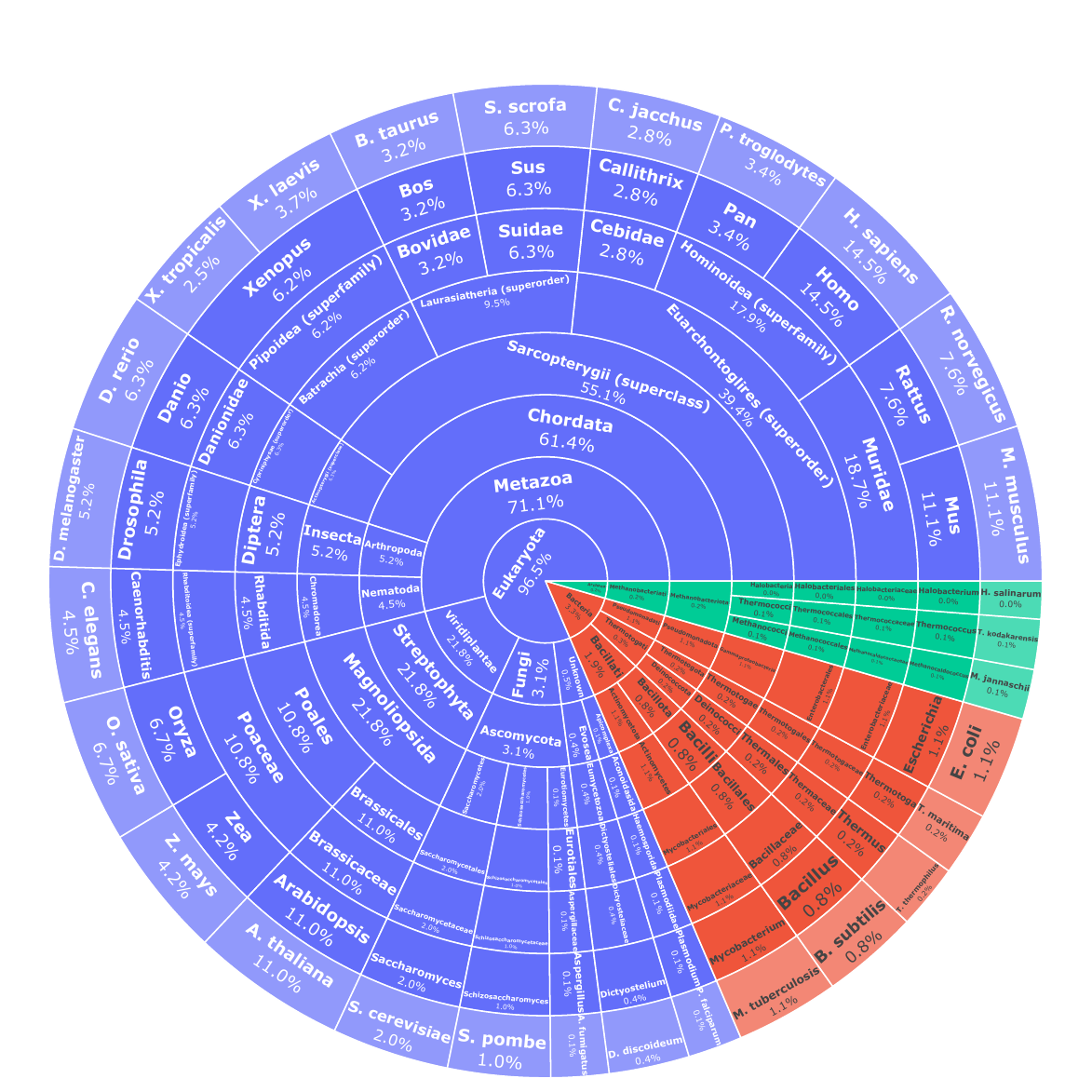}
	\caption{Proportion of different taxonomic groups in the UniProt validation set. Note that in order to make Archaea properly visible, the area for each sector is according to the log of the number of sequences in that lineage.}
	\label{fig:taxonomic_sunburst}
\end{figure*}

\Cref{fig:taxonomic_sunburst} illustrate the taxonomic lineage of the validation set,
revealing a composition dominated by Eukaryota with sparse coverage of Bacteria and
Archaea. The prevalence of Homo sapiens (human) and Mus musculus (house mouse), reflects
the anthropocentric sampling bias inherent in biological data collection.

\begin{figure*}
	\centering
	\includegraphics[width=.48\linewidth]{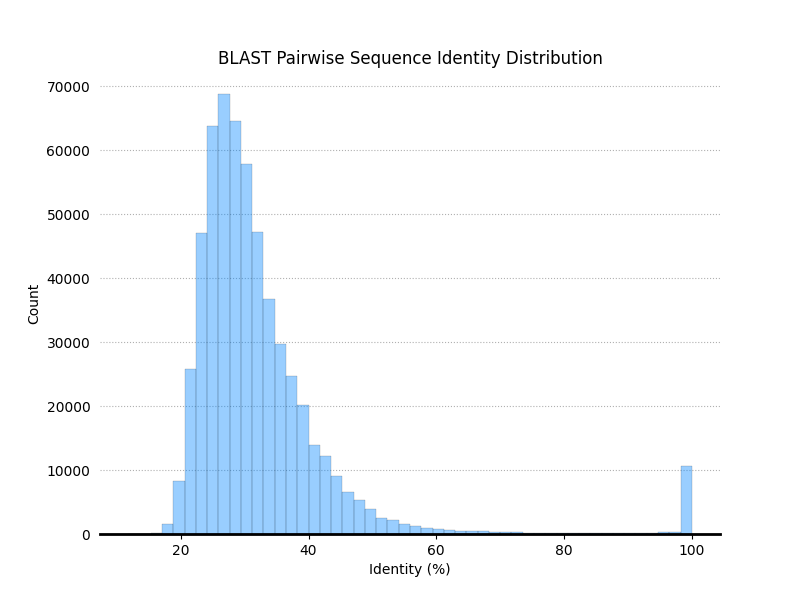}
	\includegraphics[width=.48\linewidth]{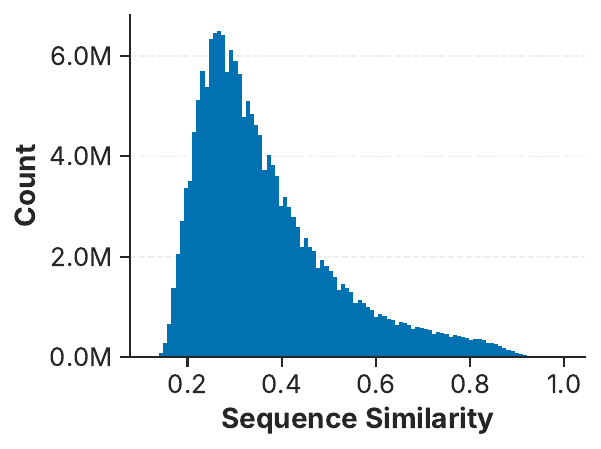}
	\caption{\textbf{Left:}Sequence similarity between different proteins in the
	validation set.
	\textbf{Right:} Histogram of closest sequence similarity for each protein in the
	training set to any protein in the validation set.
	}
	\label{fig:val_similarity}

\end{figure*}

In \Cref{fig:val_similarity} (left), we look at the pairwise sequence similarity between the
different proteins in the validation set, and see that it is quite diverse, with the
majority of sequence pairs being around 20-40\% similar i.e. in the ``protein twilight
zone''. The ``protein twilight zone'' refers to the range of low sequence identity
(typically 20--35\%) where homology inference based on sequence alignment becomes
unreliable.

In \Cref{fig:val_similarity} (right), we plot a histogram of the closest sequence
similarity for each protein in the training set to any protein in the validation set.
This shows that most proteins in the validation set have less than 40\% similarity to
any protein in the training set, indicating that the training set distribution is quite
different from the validation set distribution (even if it might contain many proteins
similar to the validation set proteins). The sequence similarities here stratified by
year are what are used in \Cref{fig:cover} to measure the distribution shifts over time.

\subsection{Stratification by Lineage}

\begin{figure*}
	\centering
	\includegraphics[width=\linewidth]{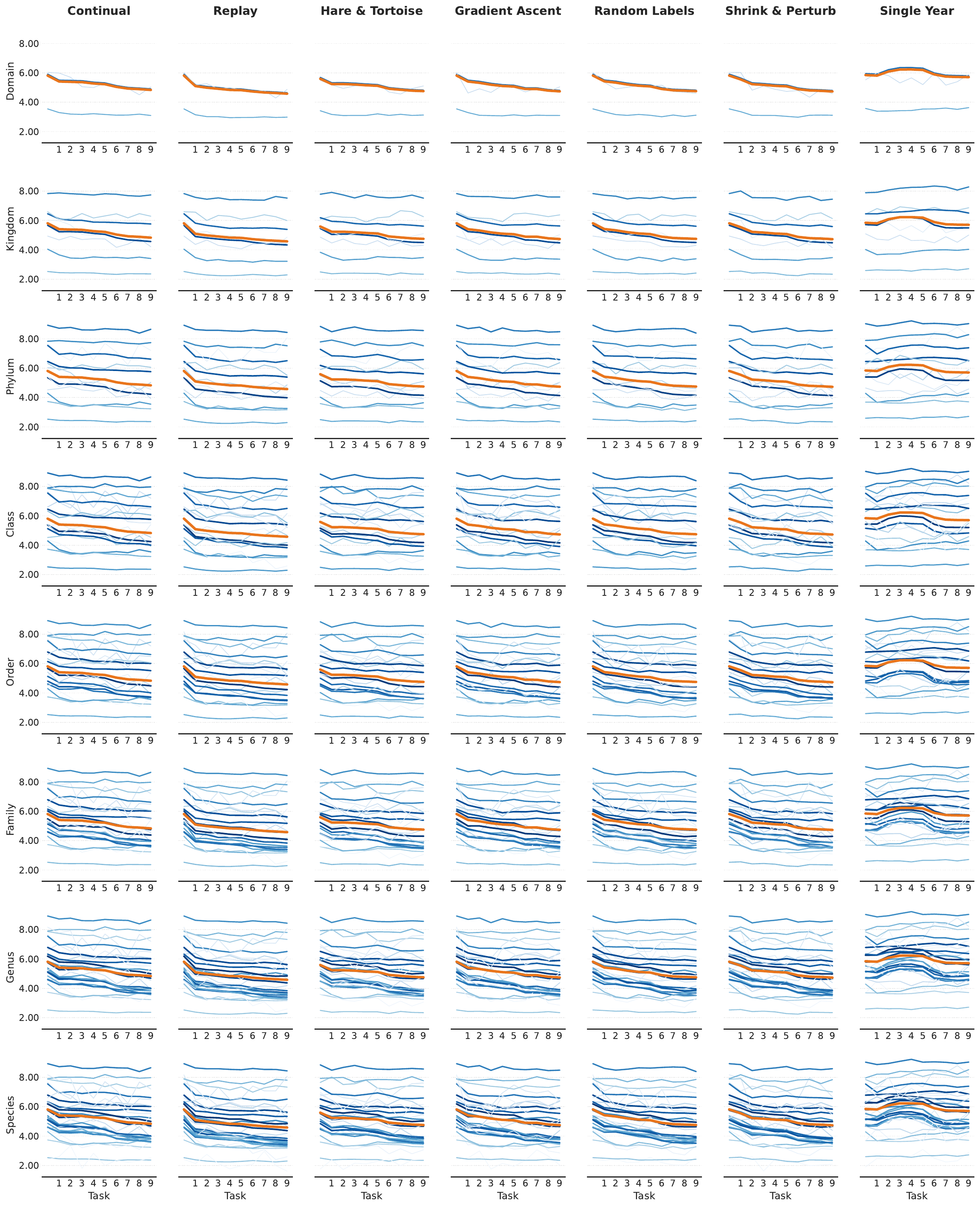}
	\caption{Perplexity on the UniProt validation set broken down by taxonomic lineage for each method. In each subplot, the mean perplexity is shown in orange, and the lines for the more common groups are shaded to be darker and thicker.
	}
	\label{fig:taxonomic_perplexity}
\end{figure*}

In this section, we present the results on the UniProt validation set for each method
stratified by the different lineages present in the dataset. We can see the results in
\Cref{fig:taxonomic_perplexity}. Notably, the mean perplexity tends to follow the
perplexity on Eukaryota and Archaea quite well, but Bacteria tends to have a much lower
perplexity. Furthermore, as we go down the taxonomic tree, it does not seem to be the
case that the model performs significantly better on the more common groups. There is a
decent amount of spread in perplexity amongst the common groups, indicating that the
models are not just memorizing the most common sequences.

\section{Fine Grained Results on ProteinGym}

\begin{figure*}[htb!]
	\centering
	\includegraphics[width=0.48\linewidth]{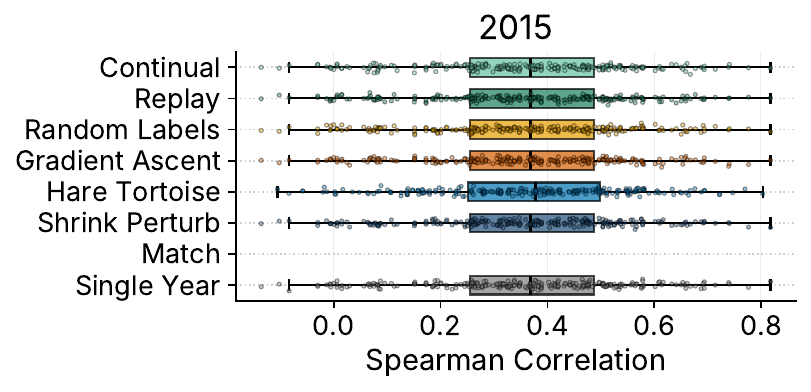}
	\includegraphics[width=0.48\linewidth]{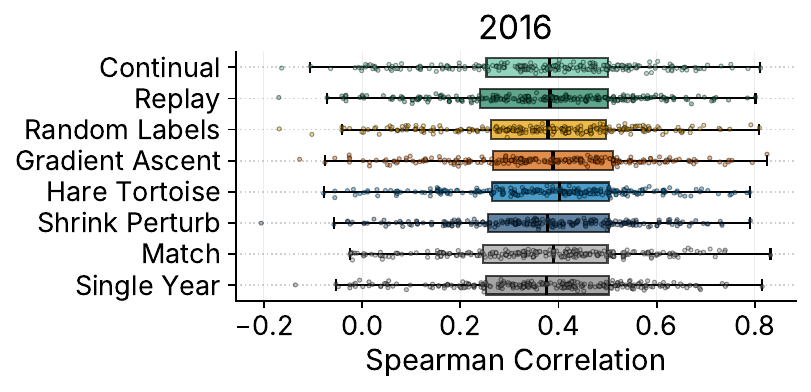}
	\includegraphics[width=0.48\linewidth]{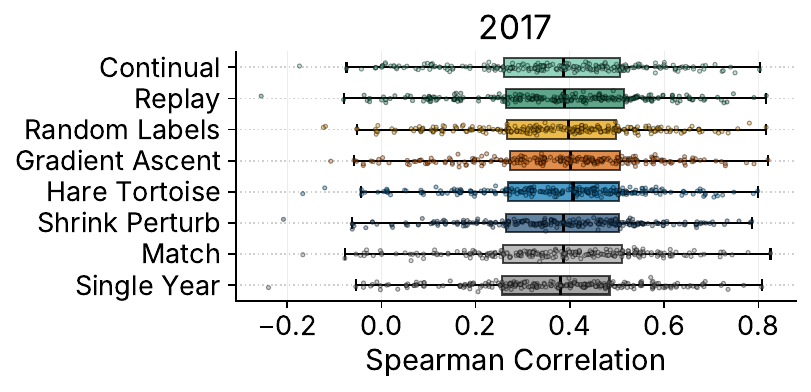}
	\includegraphics[width=0.48\linewidth]{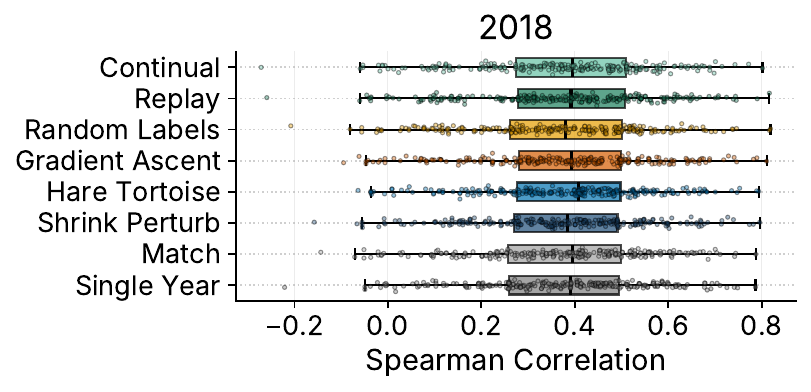}
	\includegraphics[width=0.48\linewidth]{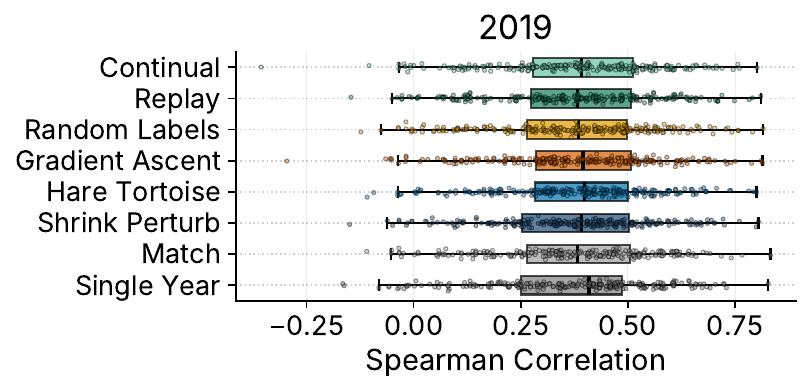}
	\includegraphics[width=0.48\linewidth]{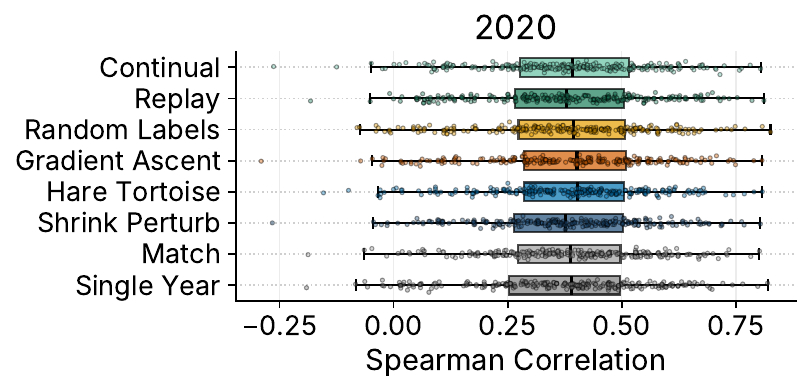}
	\includegraphics[width=0.48\linewidth]{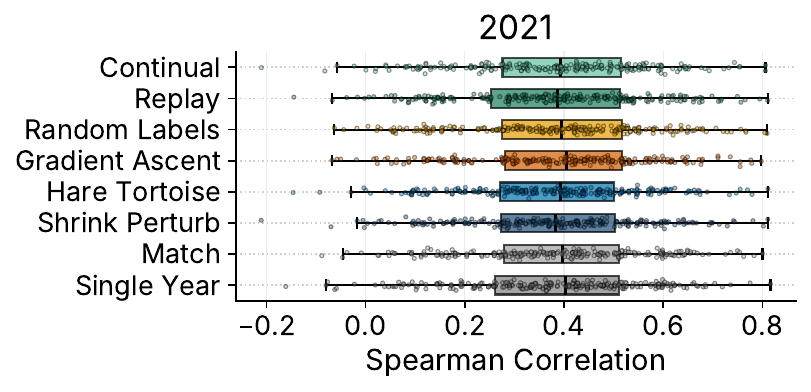}
	\includegraphics[width=0.48\linewidth]{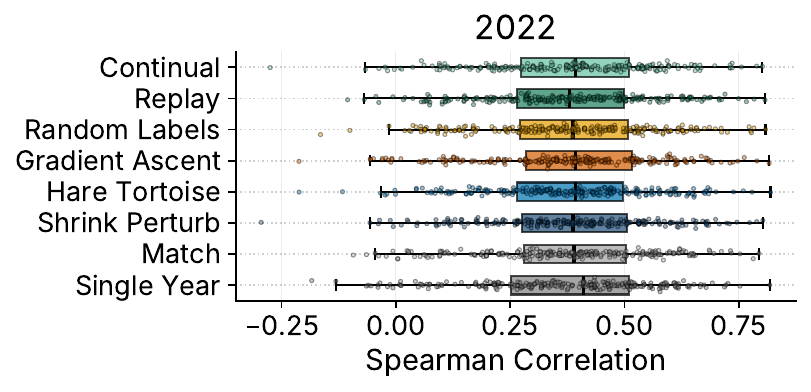}
	\includegraphics[width=0.48\linewidth]{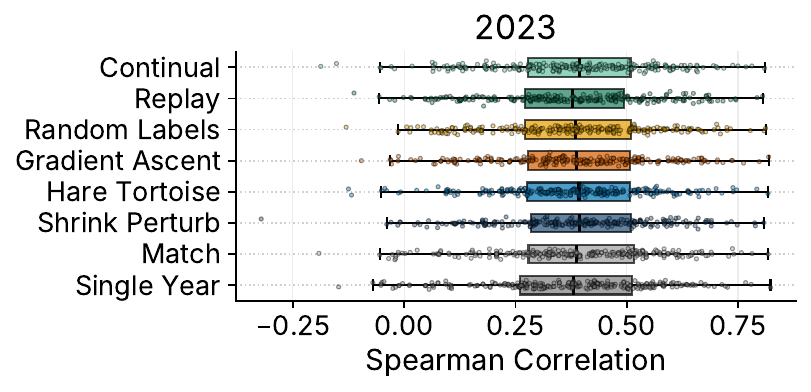}
	\includegraphics[width=0.48\linewidth]{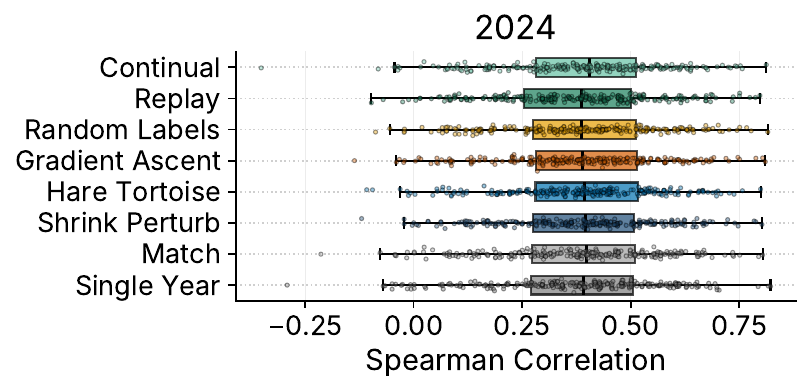}
	\caption{Distribution of Spearman correlations for each method from 2015--2024 on
	ProteinGym benchmarks. 
	}
	\label{fig:proteingym_box}
\end{figure*}

We provide the boxplots of Spearman correlations
across methods in \Cref{fig:proteingym_box}. The boxplots illustrate the distribution of
Spearman correlations for each method from 2015--2024 on ProteinGym benchmarks.
Overall, the shapes of the distributions do not vary much across years, indicating
similar performance characteristics across time.

\begin{figure*}[htb!]
	\centering
	\includegraphics[width=\linewidth]{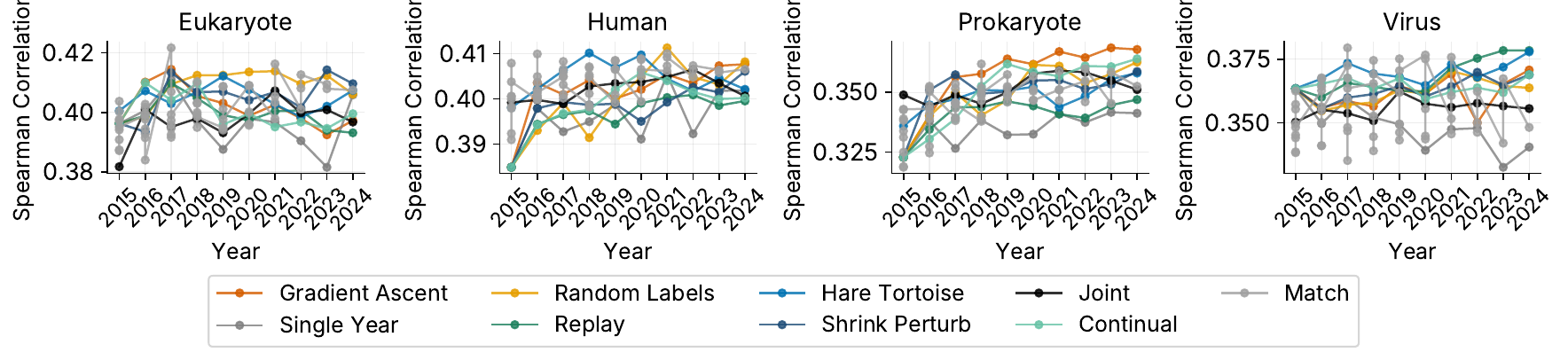}
	\caption{Spearman correlation on ProteinGym stratified by taxon.
	}
	\label{fig:proteingym_taxon}
\end{figure*}

\begin{figure*}[htb!]
	\centering
	\includegraphics[width=\linewidth]{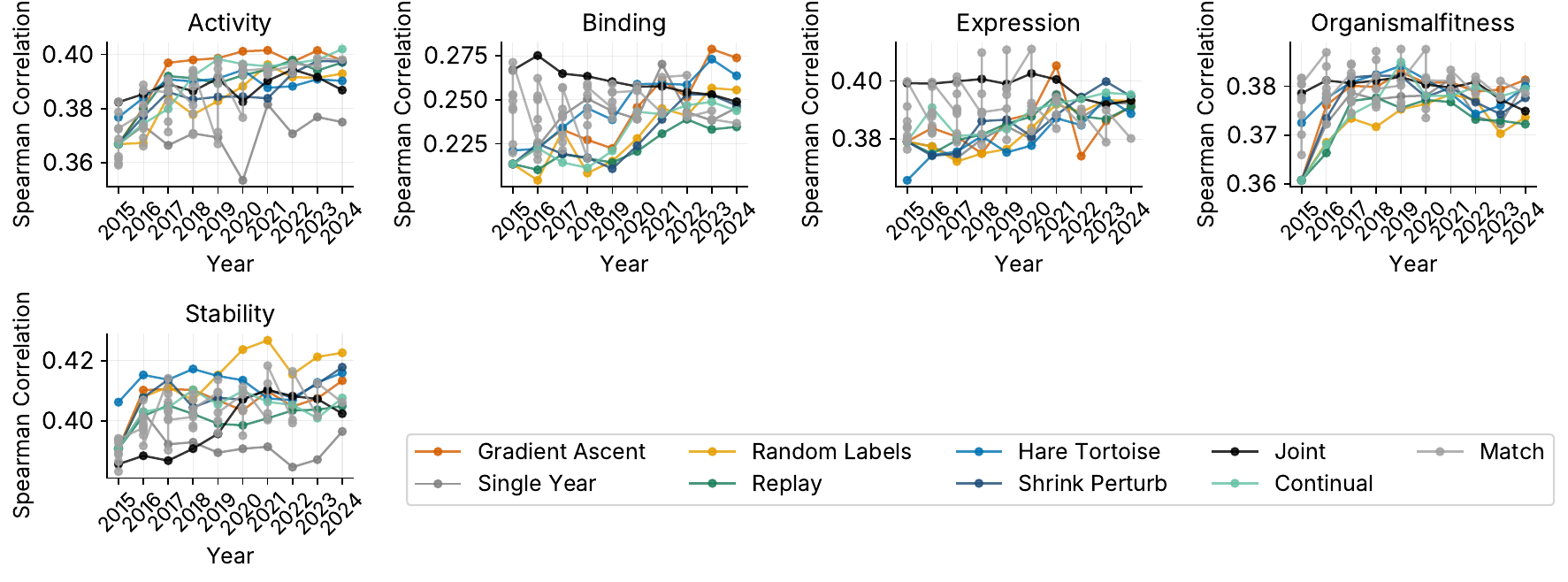}
	\caption{Spearman correlation on ProteinGym stratified by function.
	}
	\label{fig:proteingym_function}
\end{figure*}

\begin{figure*}[htb!]
	\centering
	\includegraphics[width=\linewidth]{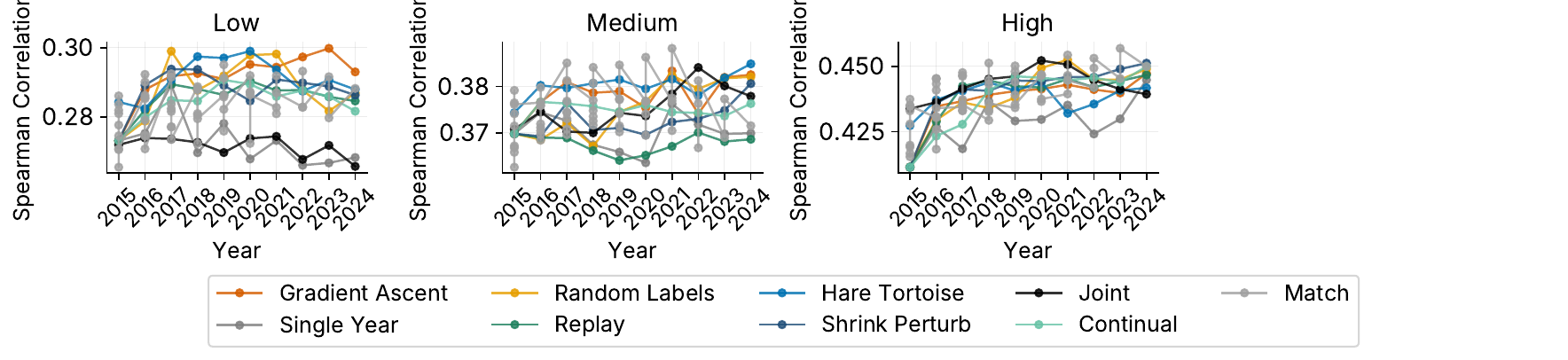}
	\caption{Spearman correlation on ProteinGym stratified by MSA Number of Effective Sequences.
	}
	\label{fig:proteingym_msa_neff}
\end{figure*}

We also provide the results on ProteinGym stratified by taxon
\Cref{fig:proteingym_taxon}, by assay function \Cref{fig:proteingym_function}, and by
Multiple Sequence Alignment Number of Effective Sequences length category (a measure of
how well studied a protein is) \Cref{fig:proteingym_msa_neff}.


\section{Fine Grained Results on PEER}
\label{appendix:PEER}

\begin{figure*}
	\centering
	\includegraphics[width=.6\linewidth]{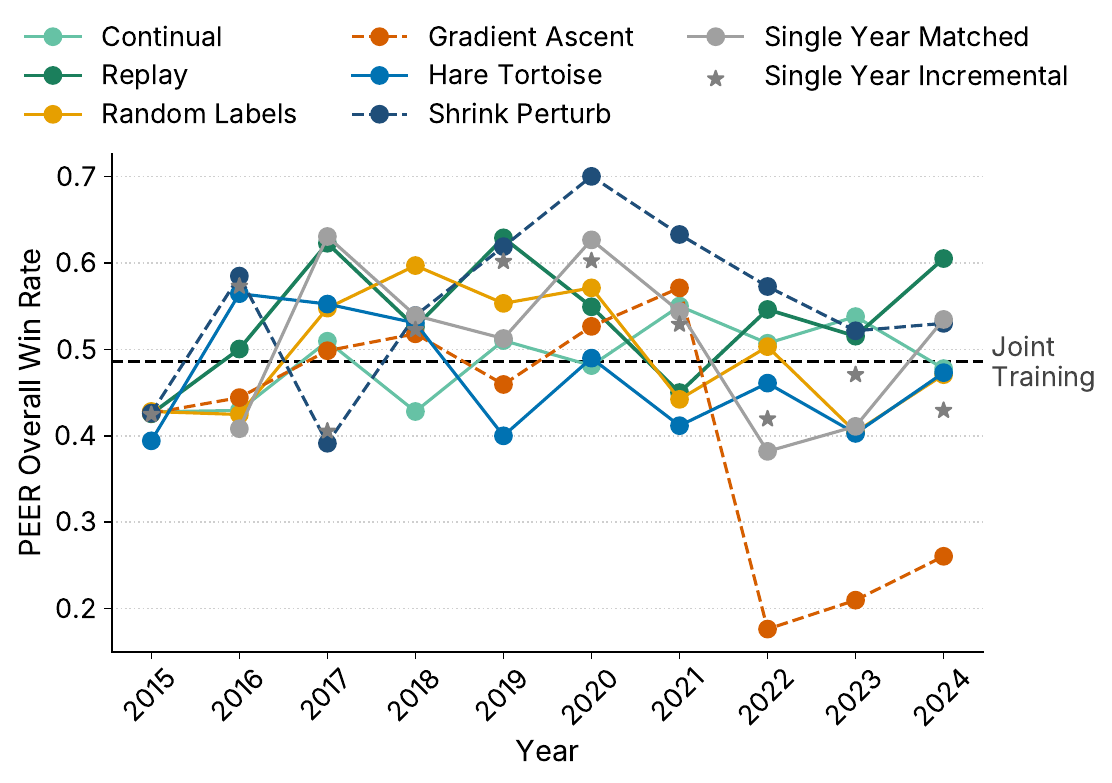}
	\caption{Win rates per year on the PEER benchmark.}
	\label{fig:peer_win_rate}
\end{figure*}

In \Cref{fig:peer_win_rate}, we show the win rates per year for each method on the PEER
benchmark. While most of the trends are difficult to discern, we see that Gradient
Ascent interestingly suffers a significant performance drop after the 2022 release, similar to what
was observed on the DGEB benchmark in \Cref{appendix:DGEB}. This might be because 2022 was one
of the larger cullings of the UniProt database, as seen in \Cref{fig:data_size}, which
could have disrupted the learning dynamics of Gradient Ascent more than other methods.

\begin{figure*}
	\centering
	\includegraphics[width=\textwidth]{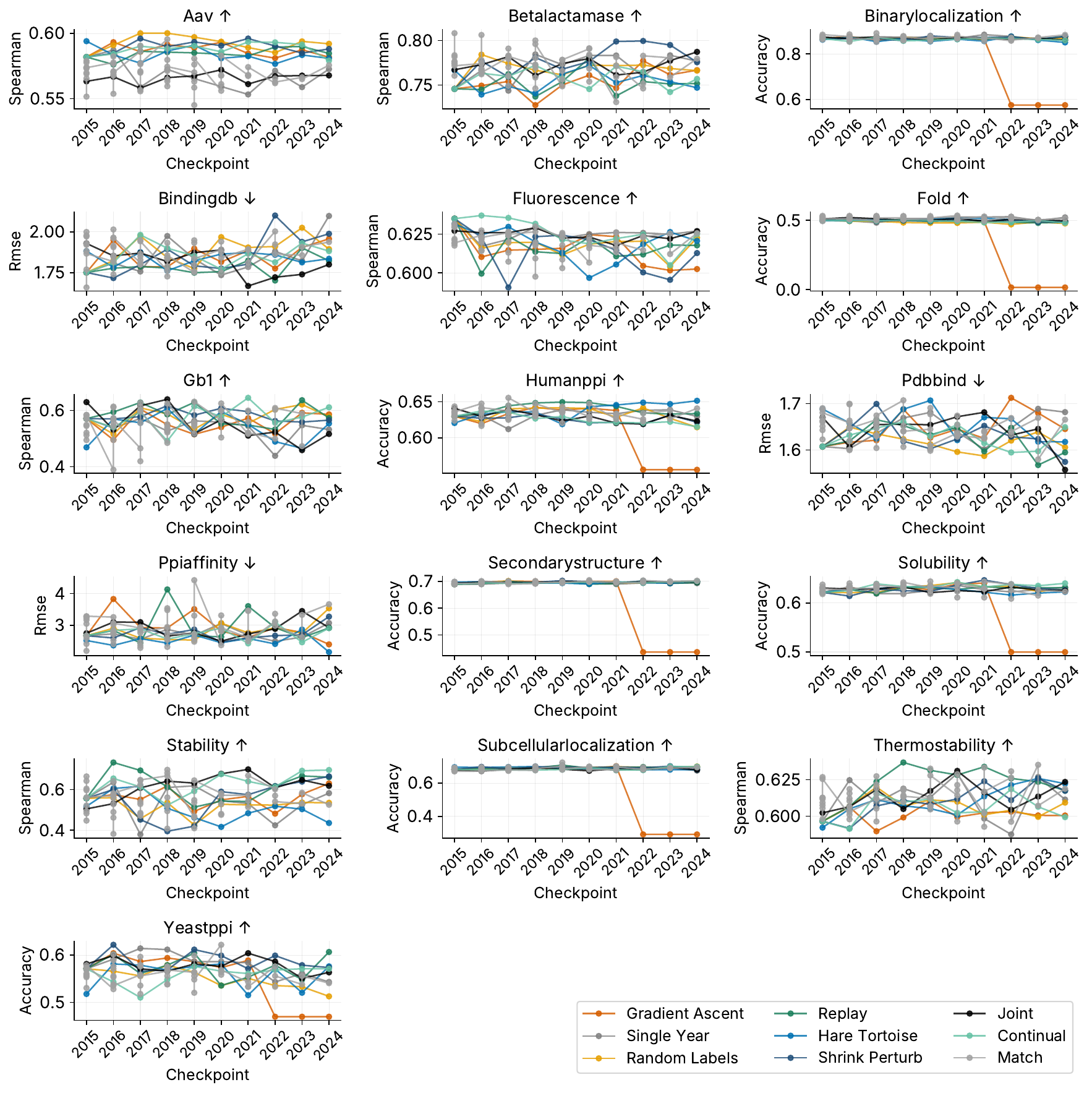}
	\caption{Full results on the PEER benchmark.}
	\label{fig:peer_full}
\end{figure*}
In \Cref{fig:peer_full}, we show the full results for each task on the PEER
benchmark.

\section{Fine Grained Results on DGEB}
\label{appendix:DGEB}

\begin{figure*}
	\centering
	\includegraphics[width=.6\linewidth]{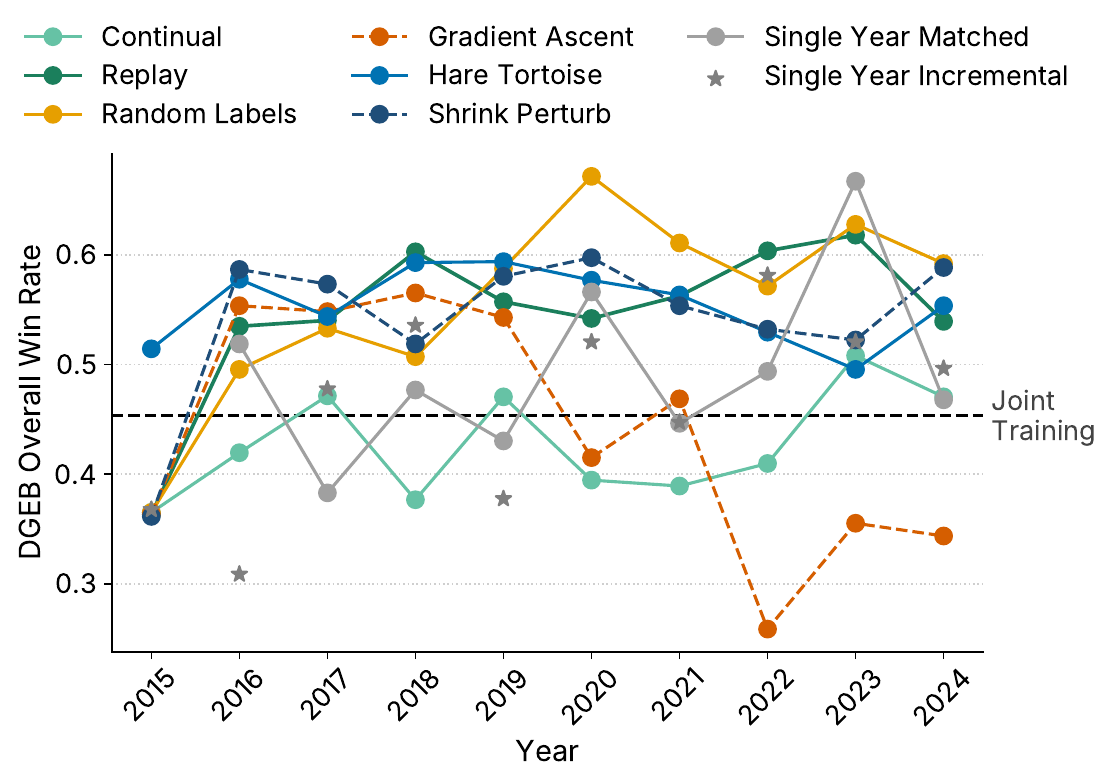}
	\caption{Win rates per year on the DGEB benchmark.}
	\label{fig:dgeb_win_rate}
\end{figure*}

In \Cref{fig:dgeb_win_rate}, we show the win rates per year for each method on the DGEB
benchmark. Similar to PEER in \Cref{appendix:PEER}, we see that Gradient Ascent suffers a
significant performance drop after the 2022 release. This might be because 2022 was one
of the larger cullings of the UniProt database, as seen in \Cref{fig:data_size}, which
could have disrupted the learning dynamics of Gradient Ascent more than other methods.

\begin{figure*}
	\centering
	\includegraphics[width=\textwidth]{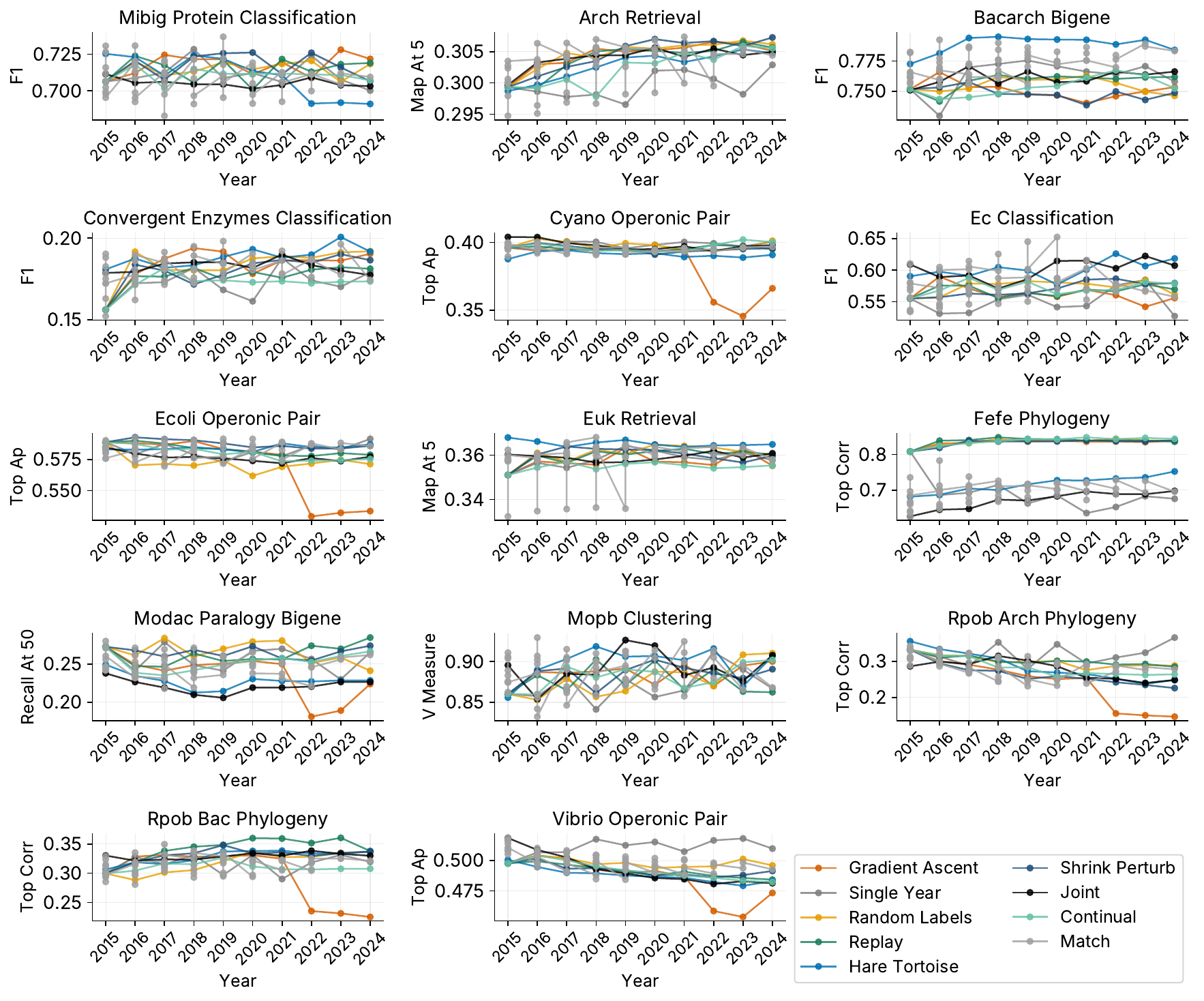}
	\caption{Full results on the DGEB benchmark.}
	\label{fig:dgeb_full}
\end{figure*}

In \Cref{fig:dgeb_full}, we show the full results for each task on the DGEB
benchmark.

\section{Further Ablations}
\label{app:ablations}

\subsection{WSD vs Cosine Learning Rate}
\label{app:lr}

\begin{figure*}
	\centering
	\includegraphics[width=.48\textwidth]{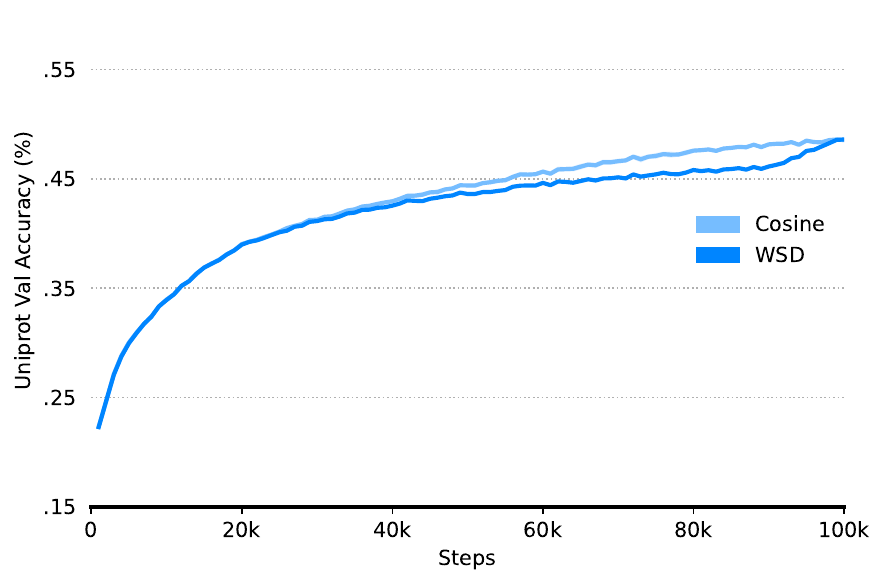}\hfill
	\includegraphics[width=.48\textwidth]{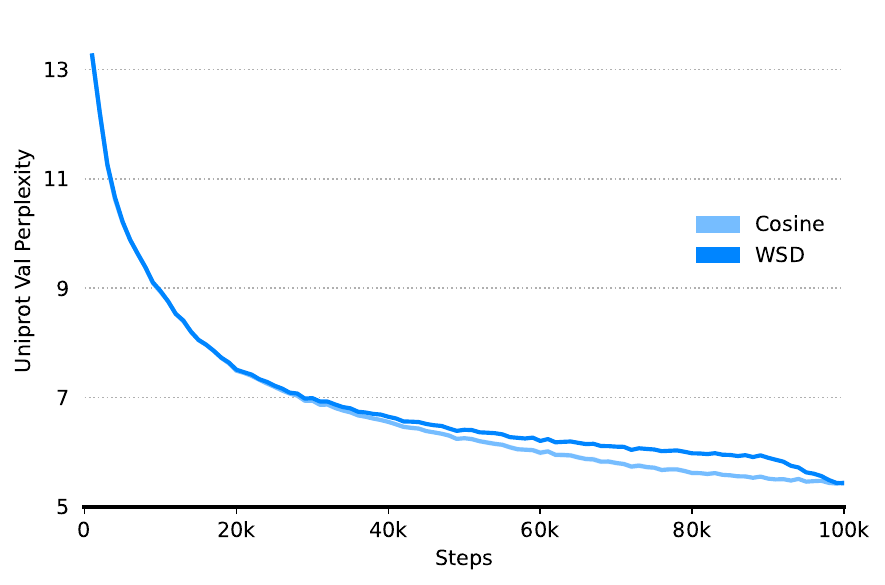}
	\caption{Both the cosine learning rate schedule and the warmup stable decay achieve approximately the same performance.}
	\label{fig:wsd_v_cos}
\end{figure*}

In this section, we clarify the learning rate schedule used by our models. Our model is based on AMPLIFY~\citep{fournierProteinLanguageModels2024}, which used a cosine learning rate schedule. Unfortunately, because the cosine learning rate schedule has a fixed span, it is unsuitable for continual training. Instead, we use the warmup-stable-decay (WSD) schedule, which has been used for continual pretraining~\citep{liModelMergingPretraining2025}. In \Cref{fig:wsd_v_cos}, we can see that after decay, the two schedules perform about equally.

In our experiments, after each decay period, we reset to the checkpoint right before the decay before moving to the next task. Thus, only 90k out of the 100k gradient steps on a task contribute to continual training, providing a good balance between the need to decay the learning rate and the ability to restart the run.

\subsection{Longer Training of a Single Year}
\label{sec:longer_training_single}

\begin{table}
	\centering
	\begin{tabular}{lr}
	\toprule
	Method & Performance \\
	\midrule
	Continual & $6.468 \pm 0.043$ \\
	\textbf{Replay} & $\mathbf{6.157 \pm 0.043}$ \\
	Shrink Perturb & $6.331 \pm 0.043$ \\
	Gradient Ascent & $6.357 \pm 0.043$ \\
	Hare Tortoise & $6.381 \pm 0.043$ \\
	Random Labels & $6.344 \pm 0.043$ \\
	Single Year (Full Run) (2015) & $6.337 \pm 0.043$ \\
	Single Year Incremental (2024) & $7.478 \pm 0.045$ \\
	Single Year Match (2024) & $6.393 \pm 0.043$ \\
	Joint & $6.571 \pm 0.044$ \\
	\bottomrule
	\end{tabular}
\caption{Results of training for an equivalent number of on a Single Year (2015)
	compared to continual training across all years.
	}
\label{tab:longer_single_year}
\end{table}
In \Cref{tab:longer_single_year}, we compare training for a longer period on a Single
Year (2015) to continual training across all years. We find that training longer on a
Single Year matches the performance of continual training. This might be because the
2015 data is particularly representative of the overall UniProt distribution, as seen in
\Cref{fig:filter}. Regardless, several of the continual learning methods outperform the
longer Single Year training, indicating that continual learning is beneficial beyond
just training for longer.

\subsection{Data Filtering Experiments with Larger Models}
\label{sec:data_filtering_at_scale}

\begin{figure*}
	\centering
	\includegraphics[width=.4\textwidth]{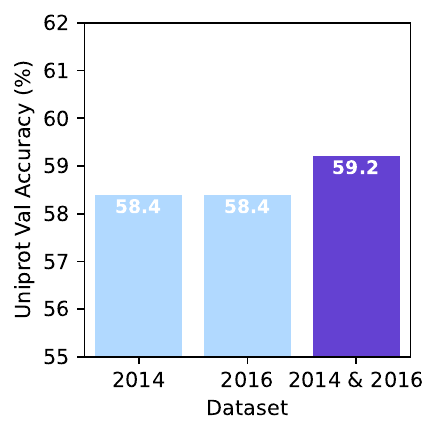}
	\caption{Data filtering experiments with a larger model (350M parameters).}
	\label{fig:filter_big}
\end{figure*}

In \Cref{fig:filter_big}, we conduct a similar data filtering experiment as in \Cref{fig:filter}, but with a larger model (350M parameters) in order to verify if the results hold at a larger scale or if they were an artifact of the smaller model potentially saturating performance. We select the intersection with the best performance in the smaller-model experiments (2014 intersected with 2016) and compare it with training on only 2014 data and only 2016 data. With the small model, we saw that training on the intersection outperformed training on either year alone. In \Cref{fig:filter_big}, we observe the same trend, indicating that this is not an artifact of model size but rather likely reflects the data quality itself.

\section{Description of Hyperparameter search and other experimental details}
\label{app:hparam}

To ensure a fair comparison, the base experimental setup is shared across all methods,
including the batch size, learning rate, and weight decay, as detailed in
\Cref{subsec:base}, and only the method-specific hyperparameters are optimized.
Since most of these algorithms have never been applied at scale, the existing literature
offers little guidance on suitable parameter ranges.
We therefore adopt an iterative, pruning-based search strategy designed to explore a
broad parameter space while rapidly discarding suboptimal settings.
For each method, we first evaluate 8 random configurations for 50k steps.
These results are used to seed a Bayesian sampler to generate 8 additional candidates
that are also evaluated at 50k steps.
From this combined pool of 16 trials, the top 4 configurations ranked by validation loss
are kept and trained for an additional 150k steps.
Finally, the optimal configuration at 200k steps is selected.

In \Cref{tab:hparam}, we describe the hyperparameter ranges and the selected value for
each hyperparameter that was searched over in our study. Each hyperparameter was sampled
independently, and we evaluated 16 trials for each method.
\begin{table*}
	\centering
	\begin{tabular}{llll}
		\toprule
		Method             & Hyperparameter          & Distribution      & Selected Value        \\
		\midrule
		Continual          & None                    &                   &                       \\
		\addlinespace
		Replay             & $\lambda_{replay}$      & Uniform(0, 1.0)   & 0.357495045651384     \\
		\addlinespace
		Hare and Tortoise  & $\lambda_{ht\_mom}$     & Uniform(.5, 1.0)  & .931247906596137      \\
		                   & $\lambda_{reset\_freq}$ & LogInt(10, 10000) & 559                   \\
		\addlinespace
		Gradient Ascent    & $\lambda_{asc}$         & Uniform(0, 1.0)   & $0.0150798214665966$  \\
		\addlinespace
		Random Labels      & $\lambda_{rand}$        & Uniform(0, 1.0)   & $0.00176366392582128$ \\
		\addlinespace
		Shrink and Perturb & $\lambda_{shrink}$      & Uniform(0, 0.9)   & $0.310430229773085$   \\
		                   & $\lambda_{noise}$       & Uniform(0, 1.0)   & $0.713412708958246$   \\
		\bottomrule
		        
	\end{tabular}
	\caption{The hyperparameter ranges and the selected hyperparameters for each method in our study.}
	\label{tab:hparam}
\end{table*}

\section{UniRef Statistics}
\label{app:data}
\begin{table*}
	\caption{The selected UniRef100 releases in our benchmark. The number of proteins
	listed are taken from the UniRef website, before we do any processing and
	deduplicating.
	}
	\label{tab:uniprot}
	\centering
    \begin{small}
	\begin{tabular}{lllll}
		\toprule
		Year & Release  & Date              & \makecell[l]{\# of Proteins} & \makecell[l]{\# of Proteins after\\ Deduplication} \\
		\midrule
		2015 & 2015\_12 & Dec. 9  & 70,511,308    & 68,944,139     \\
		2016 & 2016\_11 & Nov. 30 & 92,558,090    & 90,433,115     \\
		2017 & 2017\_12 & Dec. 20 & 128,263,573   & 125,565,575     \\
		2018 & 2018\_11 & Dec. 5  & 168,593,206   & 164,710,596     \\
		2019 & 2019\_11 & Dec. 11 & 213,522,593   & 209,351,328     \\
		2020 & 2020\_06 & Dec. 2  & 261,174,669   & 256,625,754     \\
		2021 & 2021\_04 & Nov. 17 & 280,483,851   & 275,414,005     \\
		2022 & 2022\_05 & Dec. 14 & 323,519,324   & 318,043,509     \\
		2023 & 2023\_05 & Nov. 8  & 376,564,447   & 370,850,591     \\
		2024 & 2024\_06 & Nov. 27 & 435,574,000   & 429,718,325     \\
		\bottomrule
	\end{tabular}
	\end{small}
\end{table*}
In \Cref{tab:uniprot}, we list the specific releases we used to construct our benchmark.

\end{document}